\documentclass[11pt]{article}

\usepackage[final]{acl}
\newcommand{\mci}[3]{#1~{\scriptsize[#2,\;#3]}}

\usepackage{times}
\usepackage{latexsym}

\usepackage[T1]{fontenc}

\usepackage[utf8]{inputenc}

\usepackage{microtype}

\usepackage{inconsolata}

\usepackage{graphicx}

\usepackage{booktabs}
\usepackage{amsmath}
\usepackage{amssymb}
\usepackage{multirow}
\usepackage{array}
\usepackage{enumitem}

\usepackage{tabularx}
\usepackage{dblfloatfix}
\usepackage{listings}
\newcommand{\baltrain}{\texttt{bal\_train}}
\newcommand{\balval}{\texttt{bal\_val}}
\newcommand{\testbalanced}{\texttt{test\_balanced}}
\newcommand{\xprompt}{x_{\mathrm{prompt}}}
\newcommand{\Pprompt}{\mathcal{P}}
\newcommand{\hstate}[2]{\mathbf{h}^{(#1)}_{#2}(\xprompt)}
\newcommand{\ltok}[1]{\mathrm{lasttok}_{#1}}
\newcommand{\mpool}[1]{\mathrm{meanpool}_{#1}}
\newcommand{\Lfour}{\mathrm{L4m}}

\title{Vectors Are Not Neutral: Sensitive-Information Inference from Exported LLM Representations in Summarization}

\author{
  \textbf{Weixin Liu}$^{1}$ \quad
  \textbf{Bowen Qu}$^{1}$ \quad
  \textbf{Juming Xiong}$^{1}$ \\
  \textbf{Congning Ni}$^{2}$ \quad
  \textbf{Bradley A. Malin}$^{1,2}$ \quad
  \textbf{Zhijun Yin}$^{1,2}$ \\
  $^{1}$Vanderbilt University \\
  $^{2}$Vanderbilt University Medical Center \\
  \texttt{\{weixin.liu,bowen.qu,juming.xiong\}@vanderbilt.edu} \\
  \texttt{\{congning.ni.1,b.malin,zhijun.yin\}@vumc.org}
}

\begin{document}
\maketitle

\begin{abstract}
Large language model (LLM) summarization systems may pass compact vector representations of private inputs to downstream retrieval, monitoring, audit, or analytic workflows.
Even when source documents remain access-restricted, derived vectors may be handled under different access controls and still support sensitive-information inference, creating a residual information-disclosure risk.
We study this issue in clinical discharge-summary generation as a high-stakes case study, using electronic health record (EHR)-recorded race as a controlled sensitive-label audit.
We audit two artifacts that a system might retain or expose to downstream components: the final prompt-token hidden state and the mean-pooled prompt representation.
Our results show that reducing recoverability of the case-study sensitive label from one exported artifact does not necessarily reduce recoverability from another.
As a mitigation case study, we introduce \textsc{SurfaceLoRA}, an exported-vector-targeted parameter-efficient fine-tuning method that uses a gradient-reversal discriminator attached to a designated exported vector.
Under a balanced five-way probing protocol, \textsc{SurfaceLoRA} reduces EHR-recorded race recoverability from the targeted final-token artifact toward chance while preserving summarization utility, yet recoverability remains substantially higher from untargeted pooled artifacts.
These findings show that privacy auditing and mitigation should be performed on the exact vector artifact retained or exposed to downstream components.
\end{abstract}


\section{Introduction}
\label{sec:introduction}

Large language models (LLMs) are increasingly used to summarize long, sensitive documents.
In many summarization workflows, the source text may remain access-restricted while derived vector artifacts are retained, cached, logged, indexed, or passed to downstream system components for retrieval, monitoring, auditing, or analytics~\citep{lewis2020retrieval, karpukhin2020dense, wang2021milvus, douze2025faiss, zeng2024good}.
This creates a general information-disclosure question: even when the raw texts used for summarization are protected, do derived vectors still support inference of sensitive information about the individual described by the source documents? 

This question is policy- and governance-relevant even without assuming malicious use.
A downstream component, service provider, analyst, or auditor may have access to stored vectors while lacking authorization to inspect the original text or structured sensitive attributes.
If such vectors are treated as derived artifacts that should not reveal subject-level information, then successful recovery of sensitive information from those vectors is itself a residual disclosure risk.
This concern is consistent with data-protection guidance that treats anonymization as a risk-based assessment involving residual risks such as singling out, linkability, and inference~\citep{europeanparliament2016gdpr,article29wp2014anonymisation}.
It is also supported by prior work showing that text embeddings can reveal substantial information about the underlying text~\citep{morris2023text, li2023sentence, chen2024text, huang2024transferable, chen2025algen} and that LLM-derived internal representations can create inversion or attribute-inference risks~\citep{zhu2024understanding, dong2025depth}.
Appendix~\ref{app:threat_model} provides an expanded operational illustration and threat model.

We instantiate this general problem in clinical note summarization.
Specifically, we study Brief Hospital Course (BHC) generation, where the model summarizes a hospitalization into the discharge-summary narrative describing the admission, diagnoses, treatments, clinical trajectory, and follow-up considerations~\citep{adams2021s, searle2023discharge, aali2025dataset}.
Clinical data provide a high-stakes setting in which raw notes and demographic fields are access-restricted, but derived representations may still be reused for system operations.
As a case-study sensitive attribute, we audit whether electronic health record (EHR)-recorded race can be inferred from exported vector artifacts.
We use race as a controlled example of sensitive-attribute inference rather than as the only attribute of concern; other attributes such as age, sex or gender, ethnicity, language, insurance status, and socioeconomic proxies raise analogous concerns and may require separate or multi-attribute audits.

The vector exposed by a summarization system can be defined in multiple ways.
In this study, we audit two plausible prompt-derived artifacts computed before generation: \texttt{lasttok}, the final prompt-token hidden state immediately before decoding, which serves as a compact prompt-level vector; and \texttt{meanpool}, the average of all non-padding prompt-token hidden states, which resembles pooled embeddings used for retrieval, semantic indexing, or analytics~\citep{lewis2020retrieval, karpukhin2020dense, wang2021milvus, douze2025faiss}.
Using standard post-hoc probes trained on frozen exported vectors~\citep{elazar2018adversarial, belinkov2019analysis}, we find that leakage is artifact-specific: reducing sensitive-attribute predictability on \texttt{lasttok} does not imply reduced predictability on \texttt{meanpool}.

As a mitigation case study, we introduce \textsc{SurfaceLoRA}, an exported-vector-targeted PEFT method built on LoRA and gradient reversal~\citep{hu2022lora, ganin2016domain}.
SurfaceLoRA attaches a training-time discriminator to a designated exported vector while updating only LoRA adapters and the discriminator.
Its goal is not universal sanitization, but reducing recoverability of a specified sensitive label from the specific vector artifact a system intends to retain or expose.
We evaluate frozen exported vectors with post-hoc linear and nonlinear probes that are separate from the training-time discriminator.

Our results show that \textsc{SurfaceLoRA} can drive EHR-recorded race predictability on the targeted \texttt{lasttok} artifact toward chance under a balanced five-way probing protocol while preserving BHC summarization utility.
By contrast, EHR-recorded race remains substantially more recoverable from pooled artifacts such as \texttt{meanpool}.
In addition, the utility--leakage trade-off is non-monotonic across training, motivating held-out checkpoint selection using the same exported artifact that will be retained or audited.

\noindent\textbf{Contributions.}
We make three contributions: (i) we frame exported summarization vectors as concrete audit targets for sensitive-information inference; (ii) we show that near-chance recovery from one exported vector can coexist with substantial recovery from another; and (iii) we introduce \textsc{SurfaceLoRA}, an artifact-targeted PEFT mitigation method that reduces recoverability on \texttt{lasttok} while preserving summarization utility, while pooled and multi-attribute settings require separate auditing.

\section{Related Work}
\label{sec:related_work}

Our work connects summarization, representation privacy, and adversarial mitigation.
Summarization systems are typically evaluated for generation quality, but sensitive-document summarization also raises questions about what information is preserved in intermediate artifacts that are stored, reused, or exposed.
We study this broader issue through discharge-oriented clinical summarization as a high-stakes case study.
Language models have shown strong performance on hospital-course and discharge-summary generation~\citep{adams2021s, searle2023discharge, aali2025dataset}, and MIMIC-derived corpora provide credentialed-access benchmarks built from deidentified electronic health record data~\citep{johnson2016mimic, johnson2023mimic, aali2024mimic, aali2025dataset}.
Recent clinical summarization work has emphasized verification and risk-aware evaluation beyond surface-level generation quality~\citep{asgari2025framework, chung2025verifying}.
In clinical BHC summarization, \textsc{VERI-DPO} uses claim verification to construct preferences that improve the factual support of generated summaries~\citep{liu2026veridpo}. We extend this risk-aware view from generated summaries to exported vector representations.

Representation leakage is often assessed via probing, where an attacker is trained to recover sensitive attributes from frozen representations~\citep{elazar2018adversarial, belinkov2019analysis}.
However, probe results depend on attacker capacity and the specific representation choice under audit~\citep{hewitt2019designing, pimentel2020information}.
Complementary attacks can operate on generated text or model outputs~\citep{carlini2021extracting}, while embedding-inversion attacks recover text or attributes from embedding representations~\citep{morris2023text, li2023sentence, chen2024text, huang2024transferable, chen2025algen}.
Recent studies further show that LLM-derived embeddings and internal states can expose sensitive information~\citep{zhu2024understanding, dong2025depth}.
In private dense retrieval, \citet{hua2026pointing} evaluate embedding inversion and property inference on randomized query representations. Less is known about artifact-specific leakage in summarization workflows, where different prompt-derived vectors may be retained or reused for different downstream purposes.

Mitigation strategies operate at different points in the pipeline.
Text-level approaches such as de-identification and redaction reduce disclosure in the raw text before downstream modeling~\citep{dernoncourt2017identification}, but they do not directly address what remains recoverable from transformed or embedded representations after a model processes the text.
Representation-level approaches instead aim to reduce protected-attribute recoverability from learned features.
Adversarial learning with gradient reversal trains task-useful representations while discouraging prediction of protected attributes~\citep{ganin2016domain, edwards2015censoring, madras2018learning, zhang2018mitigating, elazar2018adversarial}.
Post-hoc methods such as INLP, linear adversarial concept erasure, and LEACE remove linearly recoverable information from learned representations~\citep{ravfogel2020null, ravfogel2022linear, belrose2023leace}.
We apply a lightweight LoRA-based intervention~\citep{hu2022lora}, but our focus is deployment-oriented: auditing and mitigating the exact exported vector, then selecting checkpoints using a held-out utility--leakage trade-off.


\section{Methods}
\label{sec:method}

\paragraph{Overview.}
We instantiate the exported-vector audit in BHC generation, a clinical summarization case study where prompt-derived vectors may be cached, indexed, logged, or reused downstream~\citep{lewis2020retrieval, karpukhin2020dense, wang2021milvus, douze2025faiss, zeng2024good}.
For each example, the prompt contains the system instruction, source clinical context, and assistant generation header, but not the target BHC, generated summary, or race label.
We audit two pre-generation prompt artifacts: \texttt{lasttok}, the final prompt-token hidden state, and \texttt{meanpool}, the mean prompt-token hidden state.
Given a frozen artifact, post-hoc probes predict the five-way EHR-recorded race label from that exact vector.

\subsection{Datasets and Pre-processing}
Our primary dataset is MIMIC-IV-Ext-BHC (v1.2.0), a PhysioNet hospitalization summarization corpus derived from deidentified MIMIC-IV EHR data~\citep{johnson2023mimic,aali2024mimic}; each instance pairs a discharge note with the BHC removed and a cleaned BHC target.
We also evaluate on Discharge Me (v1.3), a BHC-adjacent MIMIC-derived task with different input construction, including chief complaint, diagnosis codes, and radiology reports~\citep{johnson2023mimic, xu2024discharge}.

For each dataset, we extract encounter-level EHR-recorded race labels and map them into five groups
(\textsc{White}, \textsc{Black}, \textsc{Hispanic}, \textsc{Asian}, \textsc{Other}), excluding \textsc{Unknown} because its assignment is ambiguous.
We split the primary dataset at the patient level into disjoint train/validation/test sets with 193{,}470/21{,}552/24{,}445 instances.
To audit race recoverability under controlled class proportions, we construct race-balanced subsets within each split by sampling an equal number of examples per race group, yielding \baltrain\ (20{,}000 total), \balval\ (2{,}500 total), and \testbalanced\ (2{,}500 total). 
These subsets are used for race-audit components, including adversary training and probe-based evaluation, while utility fine-tuning uses the full training split.
Additional preprocessing and split statistics are provided in Appendix~\ref{app:our_preprocessing}; Discharge Me details are provided in Appendix~\ref{app:dischargeme}. Because the audit subsets are class-balanced, chance accuracy is $0.20$.

\subsection{Base Summarization Model}
We fine-tune a locally deployed Llama-3.1-8B-Instruct model for BHC generation~\citep{grattafiori2024llama}.
For a given sample, let $c$ denote the source clinical context.
In the primary dataset, $c$ is the discharge-note input with the BHC section removed; it may contain detailed clinical narrative text, but it is not the full longitudinal EHR record.
We render $c$ with the system instruction and assistant generation header to obtain the input prompt $x_{\mathrm{prompt}}$.
Structured demographic labels, including EHR-recorded race, are not included in $x_{\mathrm{prompt}}$ and are used only for the adversarial and post-hoc leakage audits.
Let $y_{1:T}$ denote the target BHC tokens.
We optimize the standard token-level cross-entropy over target tokens:
\begin{equation}
\mathcal{L}_{\text{util}}(\theta)
= -\sum_{t=1}^{T} \log p_{\theta}(y_t \mid x_{\mathrm{prompt}}, y_{<t}).
\label{eq:util}
\end{equation}
where $\theta$ denotes the trainable PEFT parameters (LoRA adapters) on top of a fixed backbone; unless stated otherwise, the base LLM weights are kept fixed.
During supervised fine-tuning, the rendered prompt $x_{\mathrm{prompt}}$ is concatenated with the target BHC tokens, and prompt tokens are masked so that gradients are computed only on the target sequence.
For prompt-only representation extraction, adversarial training, and probing, we use only $x_{\mathrm{prompt}}$ without the target BHC.
We truncate $x_{\mathrm{prompt}}$ to at most 1{,}024 tokens for prompt-only passes, and truncate the full prompt--target training sequence to at most 1{,}536 tokens for supervised fine-tuning.
Results on another LLM backbone, Qwen-2.5-7B-Instruct~\citep{bai2025qwen2}, are reported in Appendix~\ref{app:qwen_backbone}.

\subsection{SurfaceLoRA: Exported-Vector-Targeted Mitigation via Adversarial PEFT}
\label{sec:slora_method}

\begin{figure*}[t]
\centering
\includegraphics[width=0.88\textwidth]{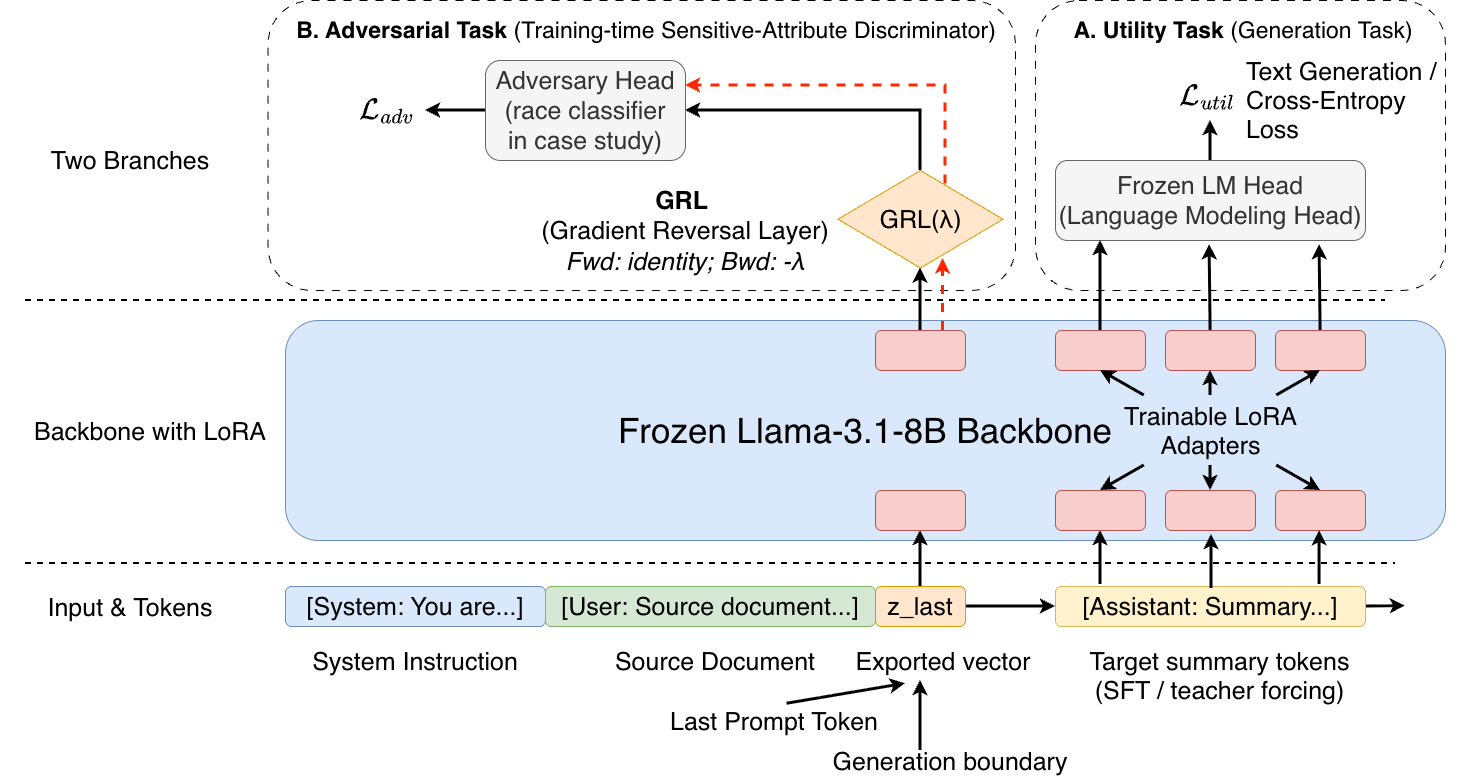}
\caption{\textbf{SurfaceLoRA mitigation training.}
SurfaceLoRA combines (A) a utility branch for summary generation with (B) an adversarial branch that attaches a gradient reversal layer (GRL)-based training-time discriminator to the exported prompt vector.
Here $z_{\mathrm{last}}$ is the exported \texttt{lasttok\_L-1} artifact at the generation boundary.
The discriminator predicts a sensitive attribute from $z_{\mathrm{last}}$; reversed gradients update LoRA adapters so the exported artifact becomes less predictive while preserving generation utility.
The discriminator is separate from the post-hoc probe attackers used for evaluation.
Only the LoRA adapters and discriminator are updated; the backbone, LM head, and decoding procedure remain fixed.
In our empirical case study, the generation task is Brief Hospital Course summarization and the audited sensitive attribute is EHR-recorded race.}
\label{fig:SurfaceLoRA_architecture}
\end{figure*}

\textsc{SurfaceLoRA} is a mitigation method, not an attribute-inference model.
It trains LoRA adapters so that a designated exported artifact remains useful for summary generation while becoming less predictive of a specified sensitive label under a training-time adversarial discriminator.
In our empirical case study, the task is BHC summarization and the sensitive label is EHR-recorded race.
The training-time discriminator is a lightweight linear classifier in the default \texttt{lasttok}-targeted experiments.
For the meanpool-targeted stress test, we also use a two-layer MLP discriminator, because pooled prompt representations may contain distributed attribute-associated signals that are not fully captured by a linear head.

For post-hoc leakage evaluation, we use two attacker classes on frozen exported representations: a multinomial logistic-regression probe as the primary linear attacker and a two-layer MLP probe as a stronger nonlinear attacker.
These probes are trained only after model training and are independent of the training-time discriminator used by \textsc{SurfaceLoRA}.  They are not used to update the LLM or LoRA adapters.
We include both linear and nonlinear probes because probe-based leakage estimates can depend on attacker capacity as well as the representation under audit~\citep{belinkov2019analysis, hewitt2019designing, pimentel2020information}.

The design principle is simple: the vector targeted during training should match the vector that will be retained, reused, or exposed.
We instantiate this principle with LoRA-based PEFT~\citep{hu2022lora} and a gradient reversal layer (GRL)~\citep{ganin2016domain}.
This follows prior work on adversarially learned and adversarially fair representations~\citep{edwards2015censoring, madras2018learning, zhang2018mitigating, elazar2018adversarial}.
At each training step, SurfaceLoRA combines a supervised fine-tuning (SFT) batch for summary generation with a prompt-only balanced batch for the adversarial sensitive-label objective.
Only LoRA adapters and the discriminator head are updated; the backbone, LM head, and decoding procedure remain fixed.

Formally, let $x_{\mathrm{prompt}}$ denote the rendered input prompt used immediately before generation.
It contains the system instruction, the source clinical context, and the assistant generation header, but does not contain the target BHC, any generated tokens, or the race label.
We run a prompt-only forward pass on $x_{\mathrm{prompt}}$ with \texttt{output\_hidden\_states=True}.
With left padding, let $i^\star = \max\{i : \texttt{attention\_mask}_i = 1\}$ be the index of the last non-padding token in $x_{\mathrm{prompt}}$.

We define the \texttt{lasttok} artifact as
\begin{equation}
f_{\theta}(x_{\mathrm{prompt}}) \;=\; \mathbf{h}^{(L-1)}_{i^\star}(x_{\mathrm{prompt}}).
\label{eq:lasttok_def_main}
\end{equation}
This vector is available at the generation boundary and can be stored as one fixed-size prompt-level artifact.

\paragraph{Layer indexing.}
We use 0-indexed transformer blocks; thus $L{-}1$ denotes the final block.

\paragraph{Alternative pooled artifact (\texttt{meanpool}).}
In addition to \texttt{lasttok}, we audit a pooled prompt representation that averages prompt-token states.
Let $\mathcal{P}$ denote the set of non-padding prompt token indices in $x_{\mathrm{prompt}}$, including the system instruction, source clinical context, and assistant generation header.
We define
\begin{equation}
f^{\texttt{meanpool}}_{\theta}(x_{\mathrm{prompt}})
\;=\;
\frac{1}{|\mathcal{P}|}\sum_{i\in\mathcal{P}} \mathbf{h}^{(L-1)}_{i}(x_{\mathrm{prompt}}).
\label{eq:meanpool_def_main}
\end{equation}

We use \texttt{meanpool} for additional representation auditing in Section~\ref{sec:redistribution}.
For the meanpool-targeted variant, the discriminator input is replaced with $f^{\texttt{meanpool}}_{\theta}(x_{\mathrm{prompt}})$ while decoding remains unchanged.
Boundary and rendering variants are detailed in Appendix~\ref{app:slora_details}.

Given a chosen exported artifact $f_{\theta}(x_{\mathrm{prompt}})$, we attach the training-time discriminator $g_{\phi}$ to predict a categorical sensitive label. In our case study, this label is the 5-way EHR-recorded race label $r \in \{1,\dots,5\}$:
\begin{equation}
\mathcal{L}_{\text{adv}}(\theta,\phi)
= \mathrm{CE}\!\left(g_{\phi}(f_{\theta}(x_{\mathrm{prompt}})),\, r\right).
\label{eq:adv}
\end{equation}

Training couples the summarization objective with exported-vector-targeted adversarial pressure through a gradient reversal layer (GRL).
At each step, we compute $\mathcal{L}_{\text{util}}$ on an SFT batch and $\mathcal{L}_{\text{adv}}$ on a prompt-only balanced batch (from \baltrain), and update parameters synchronously.
The GRL reverses gradients from $\mathcal{L}_{\text{adv}}$ flowing into the trainable adapters, yielding:
\begin{equation}
\begin{aligned}
\min_{\theta}\;& \mathcal{L}_{\text{util}}(\theta)
- \lambda\,\mathcal{L}_{\text{adv}}(\theta,\phi), \\
\min_{\phi}\;& \mathcal{L}_{\text{adv}}(\theta,\phi).
\end{aligned}
\label{eq:slora_objective}
\end{equation}
We implement this with PEFT: only LoRA adapters (and the discriminator head) are trainable, while the backbone remains fixed.
Extended implementation details, including chat-template rendering, boundary extraction, representation variants, and the training schedule, are provided in Appendix~\ref{app:slora_details}.


\subsection{Training, Decoding, and Baselines}
\label{sec:train_config}

We tune adversarial strength by sweeping $\lambda \in \{0.0, 0.02, 0.05, 0.10, 0.20, 0.50\}$ and train each configuration for 2{,}000 steps using AdamW with learning rate $2\times10^{-4}$.
Utility and adversarial objectives are optimized jointly with a fixed 1:1 utility/adversarial batch ratio.
Consistent with the setup above, prompt-only forward passes used for representation extraction, adversarial training, and probing truncate $x_{\mathrm{prompt}}$ to at most 1{,}024 tokens, while supervised fine-tuning truncates the full prompt--target sequence to at most 1{,}536 tokens.
At inference time, we provide the truncated $x_{\mathrm{prompt}}$ as input and use greedy decoding with \texttt{max\_new\_tokens=256}; this generation cap is separate from the input truncation lengths.
We keep decoding fixed across methods.
For the meanpool-targeted analysis in Section~\ref{sec:redistribution}, we use the same training, truncation, and decoding settings but attach a two-layer multi-layer perceptron (MLP) adversary to \texttt{meanpool\_L-1}; optimization details and full sweep results are provided in Appendix~\ref{app:train_details} and Appendix~\ref{app:pareto_val_points_meanpool}. The wider, coarser $\lambda$ grid for \texttt{meanpool} tests stronger gradient reversal on a representation that may combine attribute cues across prompt positions.

\paragraph{Prompt-only baselines.}
We compare \textsc{SurfaceLoRA} against two prompt-only baselines that use the same frozen instruction-tuned backbone without LoRA or adversarial training.
\textsc{Base} uses the standard clinical summarization instruction.
\textsc{Neutral} uses the same summarization instruction with an additional demographic-neutrality directive.
For both baselines, we render the source clinical context into $x_{\mathrm{prompt}}$, extract the same exported prompt representations, and evaluate them with the same post-hoc probing and ROUGE protocols as the trained models.

\subsection{Evaluation Metrics}
\label{sec:eval_axes}

To assess representation leakage, we report probe accuracy and LeakageGap (Eq.~\ref{eq:leakagegap}) for predicting the case-study sensitive label, EHR-recorded race, from the exported prompt vector.
Probe accuracy is the balanced five-way classification accuracy on \balval\ or \testbalanced, where chance accuracy is $0.20$.
LeakageGap is the absolute deviation from chance, with smaller values indicating lower recoverability under the specified post-hoc probe class.
We report these metrics for the linear and nonlinear post-hoc probes described in Section~\ref{sec:slora_method}; implementation details are provided in Appendix~\ref{app:leakage_eval_details}.

To simulate deployment-time model selection, we select checkpoints using a held-out \balval\ trade-off audit rather than defaulting to the final training step.
This validation procedure is specified before test evaluation and uses only \balval, not \testbalanced.
For the exported artifact under consideration, we first apply the LR-based validation leakage budget and then choose the feasible checkpoint with the highest validation ROUGE-L.
If no checkpoint satisfies the budget, we fall back to a Pareto-optimal low-leakage checkpoint on \balval.
The MLP probe is used only for post-hoc evaluation, not checkpoint selection.
The full checkpoint selection rule is detailed in Appendix~\ref{app:checkpoint_governance}.

To assess summarization utility, we report ROUGE-1/2/L~\citep{lin2004rouge}.
We also report two clinical utility proxies in Appendix~\ref{app:output_privacy}: BERTScore-F1~\citep{zhang2019bertscore} and concept-level overlap computed using an i2b2-2010 clinical named entity recognition (NER) model~\citep{uzuner20112010}.
Finally, to assess residual privacy risk in generated text, we report output-attacker accuracy and Macro-AUROC on \testbalanced.
We further report mention rates, defined as the percentage of generated summaries containing explicit race- or ethnicity-related terms; Appendix~\ref{app:mention_breakdown} further decomposes these terms into generic meta terms (e.g., ``race'' or ``ethnicity'') and explicit group identifiers.
Full output-attacker details are provided in Appendix~\ref{app:output_privacy}. As an additional attribute stress test, Appendix~\ref{app:gender_stress} reports a binary EHR-recorded gender audit using the same exported-vector probing protocol.

\section{Results}

\subsection{Unified Test Evaluation}
\label{sec:test_all_methods}

Table~\ref{tab:test_all_methods} provides the unified test-set evaluation on \testbalanced, including prompt-only baselines, validation-screened early checkpoints, and full-run checkpoints trained to step 2{,}000.
The table reports summarization utility with ROUGE-1/2/L and representation-level recoverability of the case-study sensitive attribute, EHR-recorded race, from the targeted \texttt{lasttok\_L-1} artifact using both LR and MLP probes.
Here, ``validation-selected'' means that checkpoint selection is performed on the same exported artifact used for the audit.
We use a pre-specified validation leakage budget of $\mathrm{LeakageGap}_{\mathrm{LR}}\leq 0.025$, i.e., within 2.5 percentage points of the balanced five-way chance baseline of $0.20$.
Among checkpoints satisfying this budget, we choose the one with the highest validation ROUGE-L; if no checkpoint satisfies the budget, we fall back to a low-leakage Pareto point.

\begin{table*}[t]
\centering
\scriptsize
\setlength{\tabcolsep}{2.4pt}
\renewcommand{\arraystretch}{0.92}

\begin{tabularx}{\textwidth}{@{}l>{\raggedright\arraybackslash}X*{7}{>{\centering\arraybackslash}c}@{}}
\toprule
\textbf{Method} & \textbf{Model/Config} &
\textbf{R-1} & \textbf{R-2} & \textbf{R-L} &
\multicolumn{2}{c}{\textbf{LR probe}} &
\multicolumn{2}{c}{\textbf{MLP probe}} \\
\cmidrule(l){6-9}
& & & & &
\textbf{Acc} & \textbf{Gap} &
\textbf{Acc} & \textbf{Gap} \\
\midrule

\multicolumn{2}{l}{Theoretical chance ($p{=}0.2$)}
& -- & -- & --
& 0.200 & 0.000
& 0.200 & 0.000 \\
\midrule

\multicolumn{9}{@{}l@{}}{\textit{Validation-screened early checkpoints (selected on \balval; evaluated on \testbalanced; bold row is selected):}} \\
\textbf{SurfaceLoRA} & \textbf{$\lambda{=}0.02$, step 1200}
& \textbf{\mci{27.48}{26.92}{28.02}}
& \textbf{\mci{7.43}{7.05}{7.78}}
& \textbf{\mci{14.54}{14.19}{14.86}}
& \textbf{\mci{0.203}{0.188}{0.219}} & \textbf{0.003}
& \textbf{\mci{0.206}{0.191}{0.222}} & \textbf{0.006} \\
SurfaceLoRA & $\lambda{=}0.20$, step 600
& \mci{22.06}{21.63}{22.46}
& \mci{5.31}{5.03}{5.57}
& \mci{12.25}{11.97}{12.52}
& \mci{0.203}{0.188}{0.219} & 0.003
& \mci{0.207}{0.192}{0.223} & 0.007 \\
SurfaceLoRA & $\lambda{=}0.05$, step 800
& \mci{23.13}{22.69}{23.55}
& \mci{5.47}{5.18}{5.74}
& \mci{12.78}{12.50}{13.06}
& \mci{0.208}{0.193}{0.224} & 0.008
& \mci{0.210}{0.194}{0.226} & 0.010 \\
\midrule

\multicolumn{9}{@{}l@{}}{\textit{Full-run checkpoints (to step 2{,}000; no early stopping):}} \\
LoRA/SFT control & $\lambda{=}0.00$
& \mci{28.04}{27.52}{28.55}
& \mci{8.08}{7.73}{8.42}
& \mci{14.90}{14.55}{15.24}
& \mci{0.230}{0.214}{0.247} & 0.030
& \mci{0.219}{0.203}{0.236} & 0.019 \\
SurfaceLoRA & $\lambda{=}0.02$
& \mci{27.05}{26.54}{27.55}
& \mci{7.31}{6.97}{7.65}
& \mci{14.34}{14.00}{14.66}
& \mci{0.227}{0.211}{0.244} & 0.027
& \mci{0.228}{0.212}{0.245} & 0.028 \\
SurfaceLoRA & $\lambda{=}0.05$
& \mci{21.76}{21.34}{22.16}
& \mci{5.34}{5.06}{5.60}
& \mci{11.65}{11.39}{11.90}
& \mci{0.234}{0.218}{0.251} & 0.034
& \mci{0.228}{0.212}{0.245} & 0.028 \\
SurfaceLoRA & $\lambda{=}0.10$
& \mci{18.95}{18.61}{19.28}
& \mci{4.44}{4.22}{4.66}
& \mci{10.27}{10.04}{10.49}
& \mci{0.232}{0.216}{0.249} & 0.032
& \mci{0.230}{0.214}{0.247} & 0.030 \\
SurfaceLoRA & $\lambda{=}0.20$
& \mci{16.76}{16.46}{17.06}
& \mci{3.79}{3.60}{3.98}
& \mci{8.91}{8.71}{9.10}
& \mci{0.236}{0.220}{0.253} & 0.036
& \mci{0.219}{0.203}{0.236} & 0.019 \\
SurfaceLoRA & $\lambda{=}0.50$
& \mci{13.64}{13.39}{13.88}
& \mci{3.30}{3.15}{3.44}
& \mci{7.24}{7.09}{7.37}
& \mci{0.228}{0.212}{0.245} & 0.028
& \mci{0.225}{0.209}{0.242} & 0.025 \\
\midrule

\multicolumn{9}{@{}l@{}}{\textit{Prompt engineering baselines (no training):}} \\
Baseline & NEUTRAL prompt
& \mci{25.37}{24.87}{25.86}
& \mci{6.11}{5.80}{6.40}
& \mci{12.47}{12.16}{12.77}
& \mci{0.234}{0.218}{0.251} & 0.034
& \mci{0.233}{0.217}{0.250} & 0.033 \\
Baseline & BASE prompt
& \mci{26.04}{25.54}{26.52}
& \mci{6.29}{5.99}{6.58}
& \mci{12.87}{12.56}{13.16}
& \mci{0.236}{0.220}{0.253} & 0.036
& \mci{0.236}{0.220}{0.253} & 0.036 \\
\bottomrule
\end{tabularx}

\caption{Test-set evaluation under a unified protocol. ROUGE is computed on \testbalanced{} ($n=2{,}500$); LR and MLP probes are fit on \baltrain{} ($n=20{,}000$) and evaluated on \testbalanced{}.
Bold marks the validation-selected checkpoint chosen by maximizing validation ROUGE-L subject to the LR-based validation leakage budget on the exported artifact ($\mathrm{LeakageGap}_{\mathrm{LR}}\leq 0.025$).
R-1/R-2/R-L denote ROUGE-1/2/L; Acc reports probe accuracy with 95\% CI; Gap is $|\mathrm{Acc}-0.2|$.
ROUGE CIs are paired bootstrap percentile intervals ($B{=}10{,}000$); probe CIs are patient-level stratified cluster bootstrap percentile intervals ($B{=}10{,}000$), resampling \texttt{subject\_id} within each race group; see Appendix~\ref{sec:statistical_methods_appendix}.}
\label{tab:test_all_methods}
\end{table*}

\subsection{Prompting and Post-hoc Removal are Insufficient}
\label{sec:prompt_posthoc_insufficient}

We first examine the prompt-only baselines defined in Section~\ref{sec:train_config}.
As shown in Table~\ref{tab:test_all_methods}, both \textsc{Base} and \textsc{Neutral} remain above chance under LR and MLP probes, indicating that instruction-level demographic neutrality is not sufficient to remove recoverable signal about the case-study sensitive attribute from the exported prompt artifact.
The \textsc{Neutral} baseline also induces output meta-term contamination: Appendix~\ref{app:mention_breakdown} shows a 46.44\% meta-term rate driven by prompt-induced wording such as ``race'' or ``racial identity'' rather than genuine demographic disclosure.

Table~\ref{tab:posthoc_lasttok} reports PCA removal, random-subspace removal, one-shot linear removal, INLP, and XCov on \texttt{lasttok\_L-1} under the same balanced five-way audit.
Appendix~\ref{sec:posthoc_baselines_appendix} shows that cached-vector post-hoc transforms mainly reduce \emph{linear} recoverability: PCA removal gives the smallest LR LeakageGap on \testbalanced\ (Acc $=0.2048$, LeakageGap $=0.0048$), but the MLP probe remains above chance (Acc $=0.2120$).
The XCov training-time decorrelation baseline also increases recoverability relative to no removal on the exported artifact.
Together, these results motivate training-time, artifact-aligned mitigation rather than relying on prompt-only constraints or linear post-processing alone.

\subsection{SurfaceLoRA Approaches the Chance Baseline on the Targeted \texttt{lasttok} Artifact}
\label{sec:main_result}

Table~\ref{tab:test_all_methods} shows that the validation-selected checkpoint ($\lambda{=}0.02$, step 1200) approaches the chance baseline on the targeted artifact while preserving utility.
It achieves LR ProbeAcc $=0.203$ (95\% CI [0.188, 0.219]) and MLP ProbeAcc $=0.206$ (95\% CI [0.191, 0.222]), with chance ($0.20$) inside both intervals.
ROUGE-L remains strong at $14.54$ (95\% CI [14.19, 14.86]), substantially above prompt-only baselines.
Thus, approximately chance-level recovery is observed for the designated exported artifact under both evaluated probe classes. At $\lambda=0$, the discriminator still trains and the GRL sends zero adversarial gradient to the LoRA adapters. At step 2{,}000, LR and MLP probes rank the $\lambda=0$ and $\lambda=0.02$ runs differently (Table~\ref{tab:test_all_methods}).

\subsection{Leakage Depends on the Exported Representation}
\label{sec:redistribution}

We next test whether mitigation transfers beyond the targeted \texttt{lasttok\_L-1} artifact.
Table~\ref{tab:rep_variant_sensitivity} shows that suppression is artifact-specific: \textsc{SurfaceLoRA} reduces \texttt{lasttok} variants to near chance (LR/MLP $\approx 0.203$--$0.208$), but \texttt{meanpool} remains substantially predictive (LR $\approx 0.316$--$0.323$; MLP $\approx 0.270$--$0.326$).
Therefore, chance-level recovery on one exported artifact does not imply model-wide sanitization; if mean-pooled embeddings are exported, the audit and adversarial target should use that representation.

We further run a meanpool-targeted sweep with an MLP discriminator and perform checkpoint selection on \balval.
Even the lowest-leakage validation checkpoint remains far above chance (LR $=0.308$, Gap $=0.108$), indicating that \texttt{meanpool} is a harder artifact to sanitize; full validation results are reported in Appendix~\ref{app:pareto_val_points_meanpool}.

\begin{table}[!t]
\centering
\footnotesize
\setlength{\tabcolsep}{2pt}
\renewcommand{\arraystretch}{0.94}
\begin{tabular}{@{}lcccccc@{}}
\toprule
\multirow{2}{*}{\textbf{Rep.}} &
\multicolumn{2}{c}{\textbf{B}} &
\multicolumn{2}{c}{\textbf{N}} &
\multicolumn{2}{c}{\textbf{S}} \\
\cmidrule(l){2-7}
& \textbf{LR} & \textbf{MLP}
& \textbf{LR} & \textbf{MLP}
& \textbf{LR} & \textbf{MLP} \\
\midrule
\texttt{lasttok\_L-1}     & 0.236 & 0.236 & 0.234 & 0.233 & \textbf{0.203} & \textbf{0.206} \\
\texttt{lasttok\_L-2}     & 0.228 & 0.223 & 0.224 & 0.230 & \textbf{0.208} & \textbf{0.206} \\
\texttt{lasttok\_L-4mean} & 0.231 & 0.217 & 0.226 & 0.228 & \textbf{0.205} & \textbf{0.208} \\
\midrule
\texttt{meanpool\_L-1}     & 0.324 & 0.270 & 0.327 & 0.266 & 0.323 & 0.305 \\
\texttt{meanpool\_L-2}     & 0.322 & 0.326 & 0.316 & 0.346 & 0.322 & 0.270 \\
\texttt{meanpool\_L-4mean} & 0.324 & 0.336 & 0.323 & 0.344 & 0.316 & 0.326 \\
\bottomrule
\end{tabular}
\caption{Representation-choice sensitivity on \testbalanced\ for the balanced five-way audit; chance accuracy is $0.20$.
\textbf{B}/\textbf{N}/\textbf{S} denote \texttt{prompt\_base}, \texttt{prompt\_neutral}, and \textsc{SurfaceLoRA} ($\lambda{=}0.02$, step 1200).
Entries are LR/MLP probe accuracies.
\texttt{L-4mean} denotes the last-four-layer mean.
Bold marks \texttt{lasttok} variants for the selected \textsc{SurfaceLoRA} checkpoint.}
\label{tab:rep_variant_sensitivity}
\end{table}
\subsection{Trade-off Fragility Requires Checkpoint Selection}
\label{sec:fragility}

The leakage--utility trade-off is non-monotonic in both training time and adversarial strength.
We therefore treat checkpoint selection as part of the deployment procedure rather than defaulting to the final training step.
This is not post-hoc selection on the test set: before test evaluation, checkpoints are selected using a pre-specified validation rule on \balval\ for the same exported representation that will be audited in deployment (Appendix~\ref{app:checkpoint_governance}).

Checkpoint choice is consequential.
For $\lambda{=}0.02$, the validation-selected step-1200 checkpoint is near chance under the MLP probe on \testbalanced\ (Acc $=0.206$, Gap $=0.006$), but continuing the same training to step 2000 increases recoverability to Acc $=0.228$ (Gap $=0.028$) (Table~\ref{tab:test_all_methods}).
This rebound suggests that adversarial pressure can temporarily reduce probe-accessible sensitive-attribute signal in the targeted representation, while later utility optimization or representation reorganization can make attribute-associated structure recoverable again.
Accordingly, deployment should audit saved checkpoints on a held-out validation split and select the checkpoint using the same exported representation and the pre-specified LR validation probe used for model selection.

\subsection{Output-Level Attribute Inference and Task Utility}
\label{sec:output_privacy_results}

Representation-level mitigation does not imply output-level sensitive-attribute invariance.
Generated summaries remain above chance under the Bio\_ClinicalBERT output attacker.
The selected checkpoint modestly improves over prompt-only baselines (Acc $0.299$ vs.\ $0.306$; Macro-AUROC $0.616$ vs.\ $0.631$), but does not approach chance.
A diagnostic attacker trained on gold BHC targets performs similarly (Acc $0.298$, Macro-AUROC $0.621$), suggesting that case-study attribute correlates are present in human-written summaries.
Mention-rate and group-wise utility analyses are reported in Appendices~\ref{app:mention_breakdown} and~\ref{app:groupwise_stability}.

\section{Discussion: Artifact-Specific Auditing for Sensitive-Information Inference}

Exported vectors from LLM-based summarization systems should be treated as artifact-specific privacy and governance surfaces.
A mitigation claim is meaningful only when it identifies both the sensitive information being audited and the exact vector artifact retained, logged, indexed, or reused.
Prior work shows that embeddings and LLM-derived representations can reveal information about underlying text or attributes~\citep{morris2023text, li2023sentence, huang2024transferable, chen2025algen, zhu2024understanding, dong2025depth}.
Our results show that recoverability can also vary across prompt-derived artifacts from the same summarization model, so auditing one vector does not justify claims about another.

\textsc{SurfaceLoRA} reduces recoverability of a chosen sensitive attribute from its designated vector artifact, but should not be interpreted as a universal sanitizer.
The leakage--utility trade-off should be evaluated separately for each audit label. The Qwen replication, Discharge Me input-construction stress test, and EHR-recorded gender audit extend the empirical evaluation within clinical summarization (Appendices~\ref{app:qwen_backbone}, \ref{app:dischargeme}, and~\ref{app:gender_stress}). In the gender audit, \texttt{lasttok} recoverability decreases while \texttt{meanpool} remains highly predictive.

Alternative exports such as \texttt{meanpool} can remain probeable even when the targeted \texttt{lasttok} artifact approaches chance-level recovery. One possible explanation is that \texttt{meanpool} aggregates race-associated cues distributed across many prompt tokens, making leakage harder to suppress with a low-rank PEFT intervention such as \textsc{SurfaceLoRA}, especially when the adversarial loss targets a single exported vector. This suggests that pooled artifacts may require dedicated mitigation, such as directly targeting \texttt{meanpool} or higher-capacity adapters.

Finally, representation-level mitigation should not be interpreted as output-level sensitive-attribute invariance.
Longer training or stronger adversarial strength do not guarantee monotonically improved invariance, so checkpoint selection should be part of deployment. In other words, it should be selected on a held-out validation trade-off set by maximizing summarization utility metric (e.g., ROUGE-L) subject to a pre-specified LeakageGap budget, with a Pareto minimum-leakage fallback.

\section{Conclusion}
We study sensitive-information inference from exported LLM representations in summarization systems. We instantiate this problem in clinical BHC summarization, using EHR-recorded race as a high-stakes case-study label for auditing whether sensitive information can be recovered from exported prompt-vector artifacts. We find that reducing recoverability from one exported vector artifact does not guarantee reduced recoverability from another. As a mitigation case study, \textsc{SurfaceLoRA} reduces recoverability from the targeted \texttt{lasttok} artifact toward chance while preserving summarization utility. However, pooled representations such as \texttt{meanpool} remain substantially predictive. We also find non-monotonic leakage--utility trade-offs, motivating checkpoint selection for the exact vector artifact that will be retained or exposed. Overall, privacy auditing for summarization representations should be artifact-specific: the vector exposed in deployment is the vector that must be audited and, when necessary, mitigated.

\section{Limitations}

Our study has several limitations that motivate future work. We study sensitive-information inference from exported LLM representations, and instantiate the problem in one empirical setting: BHC summarization with EHR-recorded race as a controlled case-study audit label. This scope enables a focused artifact-specific analysis. However, it does not establish generality across all sensitive information, summarization domains, institutions, or model architectures. \textsc{SurfaceLoRA} is designed to reduce recoverability from the representation a system plans to retain or expose, with \texttt{lasttok\_L-1} as the default targeted artifact in our experiments; accordingly, its strongest effect is expected on that artifact rather than across all internal states. Other exported artifacts, such as \texttt{meanpool} prompt embeddings, may retain sensitive information, reinforcing the need to audit the exact representation used by the system. Our evaluation uses linear and nonlinear MLP probes without exhausting all possible attackers, layers, representation combinations, or information-theoretic leakage guarantees. A further limitation is multi-attribute scalability: practical governance may require limiting inference of several sensitive factors simultaneously, such as age, sex or gender, ethnicity, language, insurance status, or socioeconomic proxies. Simply applying the same tuning process separately to each factor may be inefficient and simultaneous multi-attribute mitigation remains an important open problem. Finally, EHR-recorded race should be interpreted as an administrative category rather than a biological attribute; other sensitive or socially mediated labels will require separate definitions, baselines, probing protocols, and mitigation audits.

\section*{Ethical Considerations}
We use credentialed-access MIMIC data under the PhysioNet data use agreement (DUA) and do not release any protected text~\citep{goldberger2000physiobank, johnson2023mimic, aali2024mimic}.
Our goal is to study representation-level sensitive-information recoverability as an audit and governance risk, instantiated here with EHR-recorded race as a case-study label, not to enable attribute inference in practice.
We report coarse EHR-recorded race groupings derived from administrative registration fields.
Because race and ethnicity fields in healthcare databases can be incomplete, noisy, and institution-dependent~\citep{johnson2023accuracy}, these labels should be interpreted as recorded administrative categories rather than ground-truth identity categories or biological variables.
Our findings therefore concern the recoverability of recorded categories and their textual proxies, not biological race.
We did not send any dataset content to third-party APIs, consistent with the dataset rules.

\paragraph{Data and code availability.}
The reference implementation of \textsc{SurfaceLoRA}, including training and evaluation scripts, the main configuration, the split seed, and the prompt template, is available at \url{https://github.com/leowxvandy/SurfaceLoRA}. Credentialed PhysioNet users can prepare the clinical inputs locally as described in Appendix~\ref{app:our_preprocessing}.

\section*{Acknowledgments}
We used ChatGPT to help polish the wording of selected passages in the manuscript.

\bibliography{custom}


\appendix


\section{Extended Introduction: Threat Model, Probing, and Operational Scope}
\label{app:threat_model}

\begin{figure*}[t]
\centering
\includegraphics[width=\textwidth]{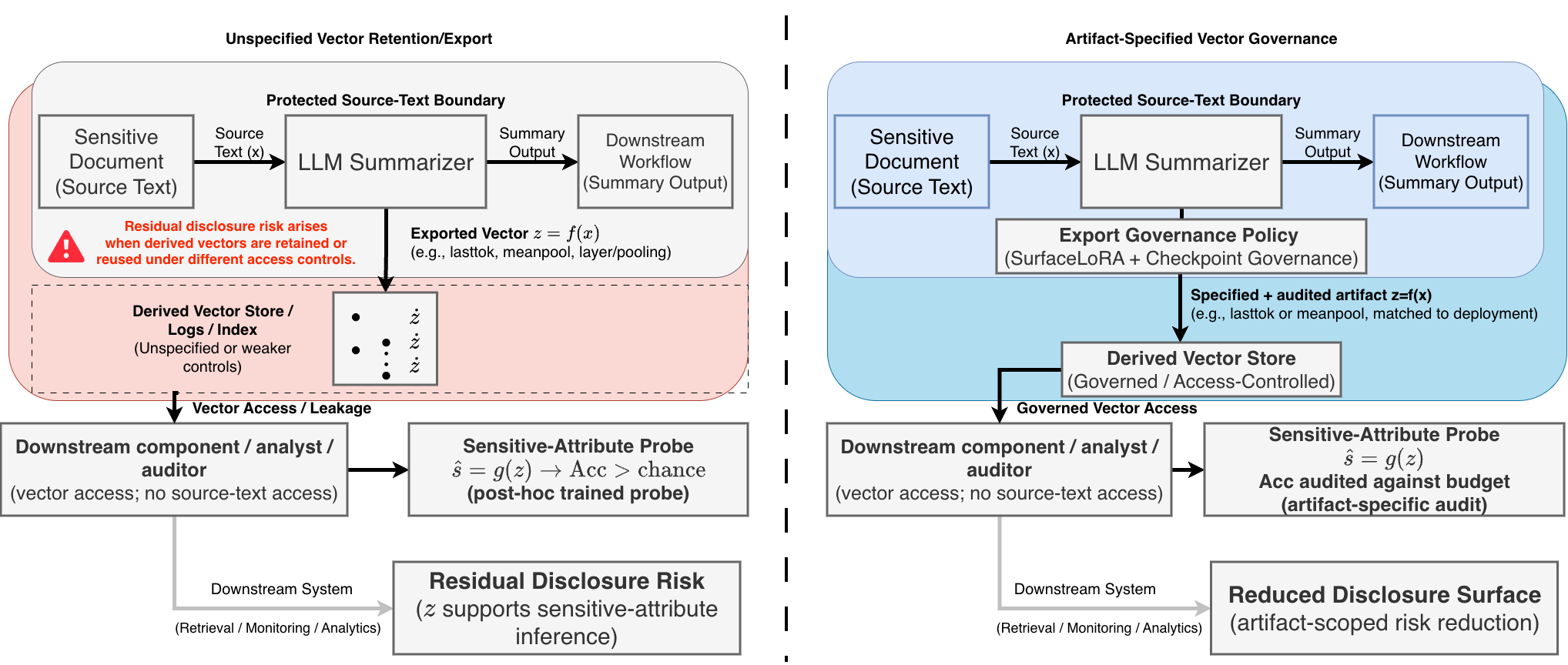}
\caption{\textbf{Vector-artifact audit for summarization systems.}
Left: a summarization pipeline may retain or reuse derived vector artifacts under controls that differ from those governing the protected source document.
A downstream component, analyst, or auditor with vector access but no source-text access may train a post-hoc probe to infer a sensitive attribute, creating a residual information-disclosure risk.
Right: an audited pipeline specifies the exact exported artifact $z=f_{\theta}(x)$, evaluates it against a leakage budget, and applies artifact-aligned mitigation and checkpoint selection.
In our empirical case study, the source documents are clinical notes and the audited sensitive attribute is EHR-recorded race.
This illustration motivates our focus on auditing the concrete exported vector artifact rather than making model-wide privacy claims.}
\label{fig:deployment_audit}
\end{figure*}

\paragraph{Artifact-specific leakage and mitigation.}
Leakage mitigation in deployed systems should be attached to the \emph{stored or reused} vector artifact.
Operationally, this is testable: sanitization on one representation (e.g., \texttt{lasttok}) does not imply sanitization on another plausible exported representation (e.g., \texttt{meanpool}).
When adversarial pressure is applied to a single exported vector (the last-token prompt state), race-associated signal can remain highly extractable from other exported representations.
For example, retrieval embeddings are often produced by a dedicated encoder; when deployments reuse the same LLM backbone for embedding, mean-pooled hidden states are a common proxy.
As a result, near-chance leakage on the targeted vector can coexist with substantial leakage on an alternative exported artifact.
This challenges the implicit assumption that suppressing leakage on a canonical token or pooling choice implies broader safety, and it can create a false sense of security if deployments export an untargeted artifact.
Accordingly, we frame leakage mitigation as attack-surface reduction on the exported vector: audit the artifact that is stored or reused, and target that same artifact during mitigation when needed.

\paragraph{What probing does (and does not) imply.}
Throughout, we use probing to quantify extractable sensitive-attribute signal under a specified attacker class~\citep{belinkov2019analysis, hewitt2019designing, pimentel2020information, elazar2018adversarial}.
High probe accuracy indicates a larger attack surface on an exposed representation, but it does not by itself establish that the generator causally uses the sensitive attribute to produce summaries.
We therefore interpret probing as a capability audit over a concrete exported vector and complement it with stronger probes and output-based attackers.
In our empirical case study, the probed attribute is EHR-recorded race.

\paragraph{Operational scope and threat model.}
Our scope is motivated by an access-boundary mismatch between protected source documents and derived representation artifacts.
A summarization system may keep the original document within a tightly controlled environment while still storing, logging, indexing, caching, or reusing derived vectors for operational workflows such as retrieval, monitoring, drift analysis, incident review, or quality analytics~\citep{lewis2020retrieval, karpukhin2020dense, wang2021milvus, douze2025faiss, zeng2024good}.
We do not assume that source documents are publicly released, nor that all systems share embeddings externally.
The risk is that a component, service provider, analyst, or auditor with access to derived artifacts under one governance boundary may not be authorized to inspect the original document or structured sensitive attributes, yet may still infer sensitive-attribute information from the vectors~\citep{morris2023text, li2023sentence, zhu2024understanding, huang2024transferable, chen2025algen, dong2025depth}.

Once such vectors are stored or reused outside the immediate generation computation, attribute inference can become feasible even when generated summaries contain no explicit demographic mentions.
We therefore study representation-level leakage and adopt attack-surface reduction as the engineering objective: reduce recoverable sensitive-label signal from a deployment-relevant representation while preserving summarization utility. In our empirical case study, the sensitive label is EHR-recorded race.

We distinguish two common vector-export settings.
First, retrieval embeddings support similarity search in RAG and dense retrieval systems~\citep{lewis2020retrieval, karpukhin2020dense}; such embeddings are commonly stored or queried through vector-indexing infrastructure~\citep{wang2021milvus, douze2025faiss}.
These are often produced by a dedicated embedding encoder, but when deployments reuse the same LLM backbone for embedding, mean-pooled hidden states are a common proxy.
Second, runtime logging features may record one vector per request for throughput-efficient monitoring, gating, auditing, caching, or drift dashboards.
SurfaceLoRA primarily targets this second setting by attaching adversarial pressure to a single prompt-level feature, the last-token state, intended to protect real-time operational logs.
We treat retrieval embeddings as a distinct attack surface: if a deployment requires mean-pooled embeddings for retrieval, the adversarial objective should be attached to that specific exported vector during training.

These distinctions align with common vector-retention scenarios.
In RAG-based assistants, embeddings may be stored in vector databases or similarity-search indexes~\citep{lewis2020retrieval, karpukhin2020dense, wang2021milvus, douze2025faiss}.
Separately, derived vectors may be used in internal, vendor-supported, or research/governance analytics (e.g., clustering, drift monitoring, cohort discovery, or quality monitoring) without giving analysts direct access to the original source text.
In such settings, the governance requirement is often asymmetric: the shared representation should preserve utility-relevant structure while limiting recoverability of sensitive attributes.
Finally, while our main audit uses EHR-recorded race labels, such supervision is plausible in clinical environments because demographics are commonly recorded as structured registration/EHR fields and may be available for authorized auditing~\citep{johnson2023accuracy}; moreover, partial disclosure, weak supervision, or linkage via auxiliary metadata can provide sufficient supervision to train attribute-inference probes.

\paragraph{Scope, setting, and boundary.}
This paper asks a targeted question: can we reduce sensitive-attribute signal in a retained or exposed internal representation while preserving summarization utility?
Our goal is not to ``wash'' generated text to eliminate all correlates of a sensitive attribute; such correlates may be task-relevant, and aggressively sanitizing outputs can harm usefulness.
Instead, we focus on reducing extractable sensitive-attribute information from a commonly reused internal feature: a prompt representation.
In our empirical case study, the sensitive label is EHR-recorded race and the summarization task is BHC generation.

We study brief hospital course (BHC) summarization in a setting motivated by common vector-use patterns in LLM systems: derived representations may be stored or reused for retrieval, monitoring, or analytics; the summarizer can be adapted locally using parameter-efficient fine-tuning (PEFT) with LoRA; and representation-level leakage can be mitigated with gradient-reversal adversarial training.
Concretely, we evaluate leakage via representation probing (linear and nonlinear probes) and utility via ROUGE plus clinical utility proxies (BERTScore and concept overlap).
Consistent with the operational framing above, we interpret improvements as reductions in what is recoverable from the chosen representation under specified attacker classes, rather than as information-theoretic removal or a claim that the generator causally uses the audited sensitive label.

\paragraph{Novelty: an artifact-specific deployment recipe.}
Our contribution is not a new adversarial objective; GRL-style adversarial learning and LoRA are established components.
Instead, we contribute an artifact-specific deployment recipe: (i) the exported vector defines the attack surface, (ii) leakage suppression is artifact-specific and should match the deployed vector, and (iii) the best utility--leakage trade-off is checkpoint-dependent rather than monotonic in training.
SurfaceLoRA instantiates this view with an exported-vector-targeted training-time discriminator under PEFT and a validation-based checkpoint selection rule.


\section{Extended Related Work}
\label{app:related_work_extended}

This work studies sensitive-attribute extractability from reusable internal representations as a practical audit issue in summarization systems.
We use clinical summarization and EHR-recorded race as a high-stakes empirical case study.
We briefly situate our approach in three areas and emphasize how our artifact-specific framing differs from much of the prior literature.

\subsection{Clinical Summarization of Discharge Narratives}
Clinical summarization aims to condense long-form notes (e.g., discharge summaries and progress notes) into concise narratives that preserve the patient trajectory, major interventions, and outcomes~\citep{adams2021s, searle2023discharge, aali2025dataset}.
Transformer-based language models have been pretrained on clinical notes and other medical text~\citep{yang2022large, chen2023meditron}.
In discharge settings, summarization is challenging due to (i) long contexts, (ii) redundant and templated sections, and (iii) the high cost of factual errors (hallucinations) and omissions, motivating careful dataset curation and evaluation beyond surface overlap metrics~\citep{asgari2025framework, chung2025verifying}.
Related work on radiology report generation uses structured evidence alignment to assess factual discrepancies between generated and reference reports~\citep{liu2026radot}.

MIMIC-based corpora have become a common testbed due to their scale and linkage to structured EHR signals~\citep{johnson2016mimic, johnson2023mimic, aali2024mimic}.
We focus on brief hospital course (BHC) generation, a clinically meaningful narrative slice of discharge documentation, and evaluate both summarization utility and recorded-attribute proxy information in model representations.

\subsection{Sensitive-Attribute Leakage and Attribute Inference via Probing}
Neural representations can encode sensitive attributes even when those attributes are not explicitly required for the task~\citep{elazar2018adversarial, zhang2020hurtful}.
A standard measurement paradigm is probing: fitting a lightweight classifier (often linear/logistic regression) on frozen representations to quantify how easily an attribute can be recovered~\citep{belinkov2019analysis}.
Probe accuracy provides an interpretable signal of extractable information, but it does not guarantee invariance to stronger adversaries or different attack surfaces, and results can depend on the representation choice and probe capacity~\citep{hewitt2019designing, pimentel2020information}.
In clinical NLP, recorded-attribute recoverability is a high-stakes instance of this broader problem because demographic fields and clinical text can be correlated in complex ways~\citep{sun2022negative}.
We adopt a controlled probing protocol on prompt representations and report deviation from chance in a balanced 5-class EHR-recorded race setting, complemented by a stronger nonlinear MLP probe and output-based attackers.

\subsection{Adversarial Invariance, Representation Sanitization, and PEFT}
Adversarial representation learning is a common strategy for reducing nuisance information in learned features.
A widely used mechanism is the gradient reversal layer (GRL), originally popularized for domain-adversarial training, which encourages representations to be predictive for the main task while uninformative for an auxiliary discriminator~\citep{ganin2016domain}.
Related ideas appear across protected-attribute leakage and privacy settings, where an adversary predicts a protected attribute and the backbone is trained (often via GRL) to reduce attribute predictability on a chosen representation~\citep{edwards2015censoring, madras2018learning}.
Beyond GRL-style objectives, other sanitization approaches include iterative nullspace projection and linear removal techniques that aim to reduce protected information from representations~\citep{ravfogel2020null, ravfogel2022linear}.
Separately, parameter-efficient fine-tuning (PEFT) adapts large pretrained models with a small number of additional parameters, enabling faster experimentation and lower storage/compute overhead.
LoRA injects low-rank updates into attention projections while keeping base weights frozen~\citep{hu2022lora}.

\subsection{Positioning and mitigation choices}
Our focus is representation-level extractability from deployment-relevant exported vectors, evaluated via probing and complemented by stronger probes and output-based attackers.
Compared with full-model adversarial training, we study a micro-adversarial setup: a single lightweight discriminator attached to a specific prompt representation under LoRA.
Beyond the method itself, we highlight a systems-level observation: the leakage--utility trade-off can be non-monotonic under training dynamics, motivating validation-based checkpoint selection as a deployment primitive.

\subsection{Post-hoc vs.\ training-time mitigation}
INLP-style and related linear removal methods can reduce linear attribute predictability via post-processing~\citep{ravfogel2020null}.
However, nonlinear extractability can persist on the same exported vector after post-hoc sanitization (Section~\ref{sec:prompt_posthoc_insufficient}), motivating training-time, artifact-targeted mitigation and validation-based checkpoint selection.


\section{Dataset Details and Access}
\label{app:data_details}

\subsection{Dataset: MIMIC-IV-Ext-BHC}
We use MIMIC-IV-Ext-BHC v1.2.0, a curated clinical-notes dataset released on PhysioNet~\citep{aali2024mimic}.
It is derived from MIMIC-IV-Note, which contains deidentified free-text clinical notes from Beth Israel Deaconess Medical Center (2008--2019) and is linkable to the broader MIMIC-IV EHR database~\citep{johnson2023mimic, aali2024mimic}.
Each record pairs (i) an \texttt{input} discharge summary with the BHC section removed and (ii) a \texttt{target} consisting of the corresponding cleaned BHC section~\citep{aali2024mimic, aali2025dataset}.
The released CSV (\texttt{mimic-iv-bhc.csv}) additionally provides \texttt{note\_id} as well as descriptive token-length metadata (\texttt{input\_tokens}, \texttt{target\_tokens}) computed with a GPT-4 tokenizer~\citep{aali2024mimic}.

The dataset authors extract the substring under the ``Brief Hospital Course'' heading using regular expressions, discard notes without a BHC section, and exclude note--BHC pairs with BHC length $<100$ characters~\citep{aali2024mimic, aali2025dataset}.
They standardize formatting (e.g., whitespace cleanup and header normalization) and retain notes containing a ``Sex'' section to support downstream subgroup analyses~\citep{aali2024mimic}.

\paragraph{Access and compliance.}
MIMIC-IV-Ext-BHC is a credentialed-access dataset distributed under PhysioNet's credentialed health data license~\citep{aali2024mimic}.
Access requires completing the required training and signing a data use agreement (DUA).
All experiments are conducted under the dataset's usage constraints; we do not release any protected text.

\begin{table}[t]
\centering
\small
\begin{tabular}{lr}
\toprule
Statistic & Value \\
\midrule
\# Note--BHC pairs & 270{,}033 \\
Input tokens (mean $\pm$ SD) & 2{,}267 $\pm$ 914 \\
Target tokens (mean $\pm$ SD) & 564 $\pm$ 410 \\
Source period & 2008--2019 \\
\bottomrule
\end{tabular}
\caption{Summary statistics of MIMIC-IV-Ext-BHC v1.2.0. Token counts are provided by the dataset and computed with a GPT-4 tokenizer for descriptive purposes only; all training and truncation use the base model's tokenizer.}
\label{tab:bhc_dataset_stats}
\end{table}

\section{Our Preprocessing and Split Construction}
\label{app:our_preprocessing}

\paragraph{Raw tables, linkage, and de-duplication.}
Starting from paired discharge-note inputs and BHC targets, we augment each example with patient- and encounter-level metadata by joining identifiers:
\texttt{note\_id} $\rightarrow$ (\texttt{subject\_id}, \texttt{hadm\_id}), then \texttt{hadm\_id} $\rightarrow$ \texttt{race}.
We de-duplicate by \texttt{note\_id} and drop rows with missing identifiers, demographics, or text fields.

\paragraph{Race normalization.}
We map raw race strings into five coarse groups, \textsc{White}, \textsc{Black}, \textsc{Hispanic}, \textsc{Asian}, and \textsc{Other}, and drop \textsc{Unknown} (e.g., unknown/declined/unable) to avoid ambiguous subgroup assignment.
Concretely, we assign \textsc{White} if the raw string contains ``WHITE'' or ``PORTUGUESE'';
\textsc{Black} if it contains ``BLACK'' or ``AFRICAN'';
\textsc{Hispanic} if it contains ``HISPANIC'' or ``LATINO'';
\textsc{Asian} if it contains ``ASIAN'';
and \textsc{Other} otherwise.
Following common practice in prior MIMIC-IV studies, we group ``WHITE'' variants (e.g., ``WHITE -- PORTUGUESE'', ``WHITE -- BRAZILIAN'') into \textsc{White} to align with dataset conventions and enable comparable subgroup evaluation~\citep{mohammed2023racial}.
We retain \textsc{Other} to preserve coverage of records that do not fall into the four named groups.

\paragraph{Text cleaning and filtering.}
We apply conservative heuristics to remove low-quality examples and reduce irrelevant long tails.
Input truncation: we truncate the input at the first occurrence of report-heavy block headers such as
\texttt{IMAGING}, \texttt{MICROBIOLOGY}, \texttt{(DISCHARGE) LABS}, \texttt{FINAL REPORT}, \texttt{RESULTS}, or \texttt{DATA};
we additionally cap inputs at 4{,}000 characters and collapse whitespace.
Length thresholds: we require the cleaned input to have at least 200 characters and the target BHC to have at least 80 characters.
Instruction-like target removal: we remove targets that resemble patient-facing discharge instructions, detected via strong phrases
(e.g., ``DISCHARGE INSTRUCTIONS'', ``FOLLOWUP INSTRUCTIONS'', ``Please call'', ``Return to the emergency room'', ``call 911'')
and weak second-person templates (e.g., ``You were admitted'') combined with action words (e.g., ``please'', ``appointment'', ``return'', ``call'', ``take'').

\paragraph{Resulting cleaned dataset.}
After cleaning, we retain 239{,}467 note--BHC pairs.
The race-group distribution is:
\textsc{White} 171{,}584 (71.65\%),
\textsc{Black} 36{,}923 (15.42\%),
\textsc{Hispanic} 13{,}426 (5.61\%),
\textsc{Asian} 7{,}873 (3.29\%),
and \textsc{Other} 9{,}661 (4.03\%).

\begin{table}[t]
\centering
\small
\begin{tabular}{lrr}
\toprule
Race group & Count & Percent \\
\midrule
White & 171{,}584 & 71.65 \\
Black & 36{,}923 & 15.42 \\
Hispanic & 13{,}426 & 5.61 \\
Asian & 7{,}873 & 3.29 \\
Other & 9{,}661 & 4.03 \\
\bottomrule
\end{tabular}
\caption{Race-group distribution in the cleaned dataset after dropping unknown or ambiguous race entries.}
\label{tab:demographics_clean}
\end{table}

\paragraph{Leakage-free patient split.}
To prevent patient-level leakage, we perform a group split by \texttt{subject\_id} using a fixed random seed.
We allocate 10\% of subjects to the test set.
From the remaining subjects, we allocate 10\% to validation and use the rest for training.
This yields 193{,}470 train (80.79\%), 21{,}552 validation (9.00\%), and 24{,}445 test (10.21\%) examples.
We verify that \texttt{subject\_id} is disjoint across train/validation/test.

\paragraph{Balanced leakage subsets.}
For controlled comparisons, we construct balanced race-stratified subsets within each split by sampling a fixed number of examples per race group
(\textsc{White}, \textsc{Black}, \textsc{Hispanic}, \textsc{Asian}, \textsc{Other}) and shuffling afterward.
We denote these subsets as \baltrain, \balval, and \testbalanced; each subset is drawn only from its corresponding train/validation/test split.
Concretely:
(i) \textbf{\testbalanced} (final reporting): 500 per class from the test split, totaling 2{,}500 examples;
(ii) \textbf{\baltrain} (stable tuning / auxiliary training): 4{,}000 per class from the training split, totaling 20{,}000 examples;
(iii) \textbf{\balval} (checkpoint selection / model selection): 500 per class from the validation split, totaling 2{,}500 examples.

\begin{table}[t]
\centering
\small
\begin{tabular}{lr}
\toprule
Split / Subset & \# Examples \\
\midrule
Cleaned total & 239{,}467 \\
Train & 193{,}470 \\
Validation & 21{,}552 \\
Test & 24{,}445 \\
\midrule
\baltrain\ (balanced) & 20{,}000 \\
\balval\ (balanced) & 2{,}500 \\
\testbalanced\ (balanced) & 2{,}500 \\
\bottomrule
\end{tabular}
\caption{Dataset sizes with patient-disjoint splits by \texttt{subject\_id}. Balanced subsets are constructed within each split for race-audit training/evaluation.}
\label{tab:splits_and_balanced}
\end{table}


\section{Extended Method Details}
\label{app:slora_details}

\subsection{Scope- and Vector-Targeted Adversarial Training}
Our adversarial intervention is deliberately scope-limited and anchored to a specific exported vector, rather than applied to all internal states or all generated outputs.
Concretely, the setup has three design constraints:
(1) a lightweight adversary, a single linear 5-way classifier head;
(2) vector-targeted pressure, adversarial gradients are applied only to a single deployment-relevant prompt vector at the generation boundary (the last non-padding prompt token state), rather than to all token states or generated tokens; and
(3) minimal disruption under PEFT, we update only the LoRA adapters (and the small adversary head) while keeping the backbone frozen.
Importantly, this scope limitation does not imply zero overhead:
under our synchronous 1:1 schedule, each step includes one SFT forward pass and one additional prompt-only forward pass to compute $\mathcal{L}_{\text{adv}}$.

\subsection{Exact \texttt{lasttok} representation definition}

\paragraph{Layer indexing.}
Let $L$ denote the number of Transformer blocks. We use 0-indexed blocks; thus $L{-}1$ denotes the final block (Python index \texttt{-1} in HuggingFace \texttt{hidden\_states}), $L{-}2$ the second-to-last, etc.

Let $c$ denote the source clinical context, and let $x_{\mathrm{prompt}}$ denote the rendered input prompt used immediately before generation.
It is obtained by applying the model-specific chat template to the system instruction and the user message containing $c$, with the assistant generation header appended.
Thus, $x_{\mathrm{prompt}}$ is the exact token sequence seen by the model immediately before it starts generating the BHC.
It contains the system instruction, source clinical context, and assistant generation header, but no target BHC tokens, generated tokens, or race label.

Unless otherwise stated, \texttt{lasttok} is extracted from this prompt-only input.
With left padding, let $i^\star = \max\{i : \texttt{attention\_mask}_i = 1\}$ denote the index of the final non-padding token in $x_{\mathrm{prompt}}$.
We define \texttt{lasttok\_L-1} as the final-block hidden state at this boundary position:
\[
\texttt{lasttok\_L-1}(x_{\mathrm{prompt}})
=
\mathbf{h}^{(L-1)}_{i^\star}(x_{\mathrm{prompt}}).
\]
Because the chat template injects role delimiters and an assistant generation header, \texttt{lasttok} represents the model state at the generation boundary, immediately before decoding begins.

\paragraph{Boundary choice (after-header vs.\ user-end).}
Our main definition corresponds to a boundary-after-header representation: the exported vector is taken after the assistant generation header is appended, immediately before decoding begins.
As a sensitivity check, we also consider a boundary-at-user-end variant by rendering the same system and user messages without the assistant generation header and taking the last non-padding token state from that sequence.
We observe the same qualitative conclusions under both boundary definitions: SurfaceLoRA drives the targeted \texttt{lasttok} artifact toward chance-level recovery at the validation-selected checkpoint, while alternative pooled representations such as \texttt{meanpool} remain substantially probeable.

\subsection{Representation variants}
Unless otherwise stated, all representation variants are computed from the same rendered prompt $x_{\mathrm{prompt}}$ defined above.
That prompt contains the system instruction, the source clinical context, and the assistant generation header, but no target BHC tokens, generated tokens, or race label.
We run a prompt-only forward pass on $x_{\mathrm{prompt}}$ with \texttt{output\_hidden\_states=True}.

Let $\mathbf{h}^{(\ell)}_i(x_{\mathrm{prompt}})\in\mathbb{R}^d$ denote the hidden state at token position $i$ from transformer block $\ell$.
With left padding, let $i^\star = \max\{i:\texttt{attention\_mask}_i=1\}$ be the index of the last non-padding token in $x_{\mathrm{prompt}}$.
We study several operational representations that could plausibly be stored or reused by downstream systems.

\smallskip
\noindent \textbf{Last-token representations (generation-boundary).}
For compact display, $\Lfour$ denotes the last-four-layer mean variant
(\texttt{L-4mean}).

{\small
\begin{align*}
\ltok{L-1}(\xprompt)
&= \hstate{L-1}{i^\star},\\
\ltok{L-2}(\xprompt)
&= \hstate{L-2}{i^\star},\\
\ltok{\Lfour}(\xprompt)
&= \frac{1}{4}\sum_{k=0}^{3}\\[-0.25em]
&\quad \hstate{L-1-k}{i^\star}.
\end{align*}
}

\smallskip
\noindent \textbf{Mean pooling over prompt tokens.}
Let $\Pprompt=\{i:\texttt{attention\_mask}_i=1\}$ be the set of non-padding token positions in $\xprompt$.
We define:

{\small
\begin{align*}
\mpool{L-1}(\xprompt)
&=
\frac{1}{|\Pprompt|}
\sum_{i\in\Pprompt}
\hstate{L-1}{i},\\
\mpool{L-2}(\xprompt)
&=
\frac{1}{|\Pprompt|}
\sum_{i\in\Pprompt}
\hstate{L-2}{i},\\
\mpool{\Lfour}(\xprompt)
&=
\frac{1}{|\Pprompt|}
\sum_{i\in\Pprompt}
\bar{\mathbf{h}}_{i}(\xprompt),\\
\bar{\mathbf{h}}_{i}(\xprompt)
&=
\frac{1}{4}
\sum_{k=0}^{3}
\hstate{L-1-k}{i}.
\end{align*}
}
\paragraph{Meanpool-targeted training.}
When meanpool-targeted training is enabled, the adversary is attached to \texttt{meanpool\_L-1} computed from the same rendered prompt $x_{\mathrm{prompt}}$.

\smallskip
\noindent \textbf{Training target vs.\ evaluation.}
SurfaceLoRA applies adversarial pressure to a \emph{single} deployment-relevant exported vector by default, typically \texttt{lasttok\_L-1}.
When a deployment exports mean-pooled embeddings, for example for retrieval or vector-store sharing, the adversarial objective should instead be attached to \texttt{meanpool\_L-1} computed from the same exported prompt artifact.

\subsection{Leakage metric (deviation from chance)}
We measure demographic extractability in a balanced 5-way race setting using probe accuracy $\text{ProbeAcc}\in[0,1]$.
Because the evaluation subsets are perfectly balanced across five classes, chance accuracy is $0.2$.
We define \texttt{LeakageGap} as the absolute deviation from chance:
\begin{equation}
\text{LeakageGap} = \left| \text{ProbeAcc} - 0.2 \right|.
\label{eq:leakagegap}
\end{equation}
For compactness, we denote LeakageGap as Gap in table headers.

\subsection{Joint training schedule (synchronous 1:1)}
At each training step, we sample (i) one SFT batch from the main training set and (ii) one prompt-only balanced leakage batch from \baltrain, using the same per-device batch size for both loaders (a strict 1:1 batch ratio).
We compute $\mathcal{L}_{\text{util}}$ on the SFT batch and $\mathcal{L}_{\text{adv}}$ on the balanced batch within the same step, then perform a single backward pass and a single optimizer step synchronously.
Equivalently, $\theta$ (LoRA adapters) is optimized to minimize $\mathcal{L}_{\text{util}}$ while maximizing $\mathcal{L}_{\text{adv}}$ via GRL, and $\phi$ (the adversary head) is optimized to minimize $\mathcal{L}_{\text{adv}}$.

\subsection{Meanpool-targeted SurfaceLoRA}
\label{app:meanpool_targeted}

SurfaceLoRA is artifact-specific: the GRL adversary is attached to the \emph{same} representation artifact that will be exported in deployment.
In our default setting, the adversary input is \texttt{lasttok\_L-1} computed from $x_{\mathrm{prompt}}$.
For \textbf{meanpool-targeted} training, we instead attach the adversary to \texttt{meanpool\_L-1}, computed by averaging the final-block hidden states over all non-padding tokens in the same rendered prompt $x_{\mathrm{prompt}}$:
{\small
\begin{equation}
\begin{aligned}
f_{\theta}^{\mathrm{mp}}(\xprompt)
&:= \mpool{L-1}(\xprompt),\\
u_{\lambda}
&:= \mathrm{GRL}_{\lambda}
\!\left(f_{\theta}^{\mathrm{mp}}(\xprompt)\right),\\
\mathcal{L}_{\mathrm{adv}}
&= \operatorname{CE}
\!\left(g_{\phi}(u_{\lambda}),\, r\right).
\end{aligned}
\label{eq:meanpool_adv}
\end{equation}
}

All other components (utility objective, LoRA parameterization, probing protocol, and checkpoint selection) remain unchanged, except that the meanpool-targeted stress test reported in Appendix~\ref{app:pareto_val_points_meanpool} uses a two-layer MLP discriminator. Thus, the targeted representation differs from the default \texttt{lasttok} setting, and the training-time adversary is strengthened for this pooled-artifact stress test.

\subsection{Parameter-efficient tuning (LoRA)}
We update the base model using LoRA adapters~\citep{hu2022lora} (rank $r{=}16$, $\alpha{=}32$, dropout $0.05$) applied to attention projection modules \texttt{q\_proj} and \texttt{v\_proj}.
All other base parameters remain frozen. In the default \texttt{lasttok}-targeted race experiments, the adversary is a single linear layer on top of the hidden state; stress-test variants using a two-layer MLP discriminator are noted explicitly.


\section{Training, Tokenization, and Inference Details}
\label{app:train_details}

\subsection{Base Model, Tokenizer, and Chat Template}
\paragraph{Base model.}
Unless otherwise stated, the main results are obtained by fine-tuning \textbf{Llama-3.1-8B-Instruct} as a local instruction-tuned causal language model (CLM)~\citep{grattafiori2024llama}.
We load the model from a local checkpoint and do not use any hosted APIs.

\paragraph{Tokenizer and chat template.}
We use the corresponding HuggingFace tokenizer (\texttt{AutoTokenizer}, fast tokenizer enabled) and the model-provided chat template to construct prompts.
For each example, we create a two-message list with roles \texttt{system} and \texttt{user}, and render it via
\texttt{tokenizer.apply\_chat\_template(messages, add\_generation\_prompt=True)}.
Using the tokenizer-provided template ensures that special tokens and role delimiters follow the exact Llama-3.1-Instruct formatting specified by the tokenizer configuration, which matters because hidden states depend on these tokens.
We use left padding for both training-time batching and evaluation-time generation.
If the tokenizer has no explicit pad token, we set \texttt{pad\_token = eos\_token}.

\paragraph{Truncation.}
Prompt-only inputs are truncated to at most 1{,}024 tokens by keeping the most recent tokens (left truncation of the oldest context).
For SFT (prompt + target), sequences are truncated to at most 1{,}536 tokens using the same left-truncation rule to fit GPU memory constraints.

\subsection{Training Configuration}
\paragraph{Sweep.}
For \texttt{lasttok}-targeted SurfaceLoRA, we sweep GRL coefficients
$\lambda \in \{0.0, 0.02, 0.05, 0.10, 0.20, 0.50\}$ with a fixed random seed.
Each run trains for 2{,}000 steps using AdamW with learning rate $\mathrm{lr}=2\times10^{-4}$.
For the meanpool-targeted sweep, we use $\lambda \in \{0.0, 0.05, 0.10, 0.30, 1.0\}$ and evaluate checkpoints up to step 1{,}400 under the same checkpoint selection procedure. This sweep uses a two-layer MLP discriminator attached to \texttt{meanpool\_L-1}.

\paragraph{Optimization.}
We use per-device batch size 4 and gradient accumulation 32, resulting in an effective batch size of 128.
We apply gradient clipping with max norm 1.0.
Unless otherwise stated, AdamW uses PyTorch defaults (betas $(0.9, 0.999)$, $\epsilon=10^{-8}$, and weight decay $0.01$).
We do not use a learning-rate scheduler or warmup (fixed learning rate for all steps).

\paragraph{Joint optimization schedule (simultaneous update; 1:1 batch ratio).}
We optimize the utility objective and the adversarial objective jointly (not alternating separate steps).
At each training step, we sample (i) one SFT batch from the main training set and (ii) one prompt-only balanced leakage batch from \baltrain, using the same per-device batch size for both loaders (a strict 1:1 batch ratio).
We compute $\mathcal{L}_{\text{util}}$ on the SFT batch and $\mathcal{L}_{\text{adv}}$ on the balanced batch within the same step, and aggregate them as:
\begin{equation}
\begin{aligned}
u_{\lambda}
&:= \mathrm{GRL}_{\lambda}
\!\left(f_{\theta}(\xprompt)\right),\\
\mathcal{L}_{\mathrm{total}}
&=
\mathcal{L}_{\mathrm{util}}
+
\operatorname{CE}
\!\left(g_{\phi}(u_{\lambda}),\, r\right).
\end{aligned}
\label{eq:train_total_schedule}
\end{equation}
followed by a single backward pass and a single optimizer step that updates all trainable parameters synchronously.

\paragraph{Equivalent optimization view.}
With GRL, the synchronous update is equivalent to
\begin{align*}
\min_{\theta}\quad
& \mathcal{L}_{\mathrm{util}}(\theta)
- \lambda\,\mathbb{E}[\mathcal{L}_{\mathrm{adv}}],\\
\min_{\phi}\quad
& \mathbb{E}[\mathcal{L}_{\mathrm{adv}}].
\end{align*}
Here, $\mathcal{L}_{\mathrm{adv}}=\mathcal{L}_{\mathrm{adv}}(\theta,\phi)$.
The GRL reverses and scales gradients flowing into $f_{\theta}(\xprompt)$ and thus into $\theta$, while gradients with respect to $\phi$ are unchanged.

\paragraph{Optimizer sharing and $\lambda$ schedule.}
We use a single shared AdamW optimizer over the union of trainable LoRA parameters and adversary-head parameters (no separate optimizers or time-sharing updates).
Unless otherwise stated, $\lambda$ is held constant throughout training (no warm-up or annealing), and we sweep $\lambda$ across runs to control adversarial strength.

\paragraph{Precision and memory.}
On GPU, we run in bfloat16 precision by loading the base model with \texttt{torch\_dtype=bfloat16}.
We enable gradient checkpointing to reduce memory usage.
We allow TF32 for CUDA matmul/cuDNN to improve throughput where supported.
We attempt to enable FlashAttention-2 via \texttt{attn\_implementation=flash\_attention\_2} when available~\citep{dao2023flashattention}; otherwise we fall back to the default attention implementation.
We do not use DeepSpeed/ZeRO; each sweep run trains in a single GPU process.

\paragraph{Sequence lengths and generation.}
We set the maximum SFT sequence length ($x_{\mathrm{prompt}}$ + target BHC tokens) to 1{,}536 tokens and the maximum prompt-only length for $x_{\mathrm{prompt}}$ to 1{,}024 tokens.
At evaluation time, we use greedy decoding (no sampling) and cap the output to 256 new tokens.

\paragraph{Inference settings (fixed across all methods).}
For all reported ROUGE results, we use greedy decoding with \texttt{do\_sample=False}, \texttt{max\_new\_tokens=256}, and the default repetition penalty.
We keep the decoding configuration identical across the baseline and all SurfaceLoRA runs to ensure comparability.

\paragraph{Evaluation stability (out-of-memory (OOM) avoidance).}
To reduce evaluation-time OOM risk, we generate with small micro-batches (e.g., 8 prompts per batch) and left padding.
To mitigate CUDA memory fragmentation, we set \texttt{PYTORCH\_CUDA\_ALLOC\_CONF} to \texttt{expandable\_segments:True}.


\section{Representation Leakage Evaluation Details}
\label{app:leakage_eval_details}

\subsection{Representation Leakage Probe Protocol}
We quantify EHR-recorded race extractability from prompt representations using probe classifiers.
For each example, we render the source clinical context into $x_{\mathrm{prompt}}$ and extract $f_{\theta}(x_{\mathrm{prompt}})$, the exported prompt representation under the specified representation choice. Under the default setting, this is the \texttt{lasttok\_L-1} artifact. We then fit a 5-way multinomial logistic regression probe on \baltrain{} (20{,}000; balanced).
We evaluate the trained probe on \balval{} (2{,}500; balanced) for model selection and on \testbalanced{} (2{,}500; balanced) for final reporting.
Let $\text{ProbeAcc}\in[0,1]$ denote probe accuracy on balanced 5-way subsets; chance accuracy is therefore $0.2$.
We report \texttt{LeakageGap} using the deviation-from-chance definition in Eq.~\ref{eq:leakagegap}.

\paragraph{Linear probe implementation.}
Unless otherwise stated, the linear probe is a 5-way multinomial logistic regression trained on \baltrain{} with fixed hyperparameters:
\texttt{solver=saga}, \texttt{penalty=l2}, \texttt{multi\_class=multinomial}, \texttt{max\_iter=800}, \texttt{tol=1e-3}, and \texttt{n\_jobs=16}, using a fixed random seed for reproducibility.
We then evaluate the probe on \balval{} or \testbalanced{} and report probe accuracy along with $\texttt{LeakageGap}=|\texttt{ProbeAcc}-0.2|$.

\paragraph{Probed representations.}
Unless otherwise stated, the main tables probe the default exported artifact, \texttt{lasttok\_L-1}.
For analysis, we additionally probe alternative operational representations (mean pooling over prompt tokens and last-4-layer averaging; see Section~\ref{sec:redistribution} and Appendix~\ref{app:slora_details}) to assess whether leakage reduction is global or confined to the targeted artifact.

\paragraph{Stronger probe: MLP classifier.}
To address the possibility that a linear probe underestimates extractable demographic information, we additionally evaluate a 2-layer MLP probe (hidden dimension 128, ReLU, dropout 0.3).
We train the MLP on \baltrain{} using AdamW (learning rate $10^{-3}$, batch size 128) and select the best checkpoint on \balval{} via early stopping (patience 5 epochs; maximum 50 epochs).
This probe is strictly more expressive than logistic regression and can capture nonlinear structure in the representation space.


\section{Checkpoint Selection Policy}
\label{app:checkpoint_governance}

\paragraph{Model selection is artifact-specific.}
We select deployable checkpoints based on validation performance \emph{on the exported artifact} $z$.
For a given exported representation $z$ (e.g., \texttt{lasttok\_L-1} or \texttt{meanpool\_L-1}), checkpoint selection uses the logistic-regression probe as the pre-specified validation attacker. We first apply the LR-based validation leakage budget and then maximize validation ROUGE-L among feasible checkpoints.
If no checkpoint satisfies the budget, we select a Pareto-optimal checkpoint with minimum $\textit{LeakageGap}_{\mathrm{LR}}(c;z)$.
We additionally report Pareto-optimal checkpoints under minimizing $\textit{LeakageGap}_{\mathrm{LR}}(c;z)$ and maximizing ROUGE-L$(c)$.
Operationally, this treats the \balval{} audit as a deployment decision rule: the deployable model is the selected checkpoint under the \emph{same exported representation} that the system uses.

\begin{center}
\fbox{%
\begin{minipage}{0.97\linewidth}
\small
Checkpoint selection as a deployment primitive (\emph{validation-only; artifact-specific}).\\
Input: checkpoints saved during training.\\
Choose exported representation $z$ (e.g., \texttt{lasttok\_L-1} or \texttt{meanpool\_L-1}, computed from $x_{\mathrm{prompt}}$ unless otherwise stated).\\
Audit: for each checkpoint $c$, compute $\textit{LeakageGap}_{\mathrm{LR}}(c;z)$ using an LR probe fit on \baltrain\ with representation $z$ and evaluated on \balval\ with the same representation; also compute validation \textit{ROUGE-L}$(c)$ on \balval.\\
Constraint: retain checkpoints with $\textit{LeakageGap}_{\mathrm{LR}}(c;z) \le \epsilon$.\\
Selection: among feasible checkpoints, choose the checkpoint with the highest validation \textit{ROUGE-L}.\\
Fallback: if no checkpoint is feasible under $z$, choose a Pareto-optimal checkpoint with minimum validation $\textit{LeakageGap}_{\mathrm{LR}}(c;z)$.\\
Test use: evaluate the selected checkpoint once on \testbalanced.
\end{minipage}}
\end{center}

We set $\epsilon=0.025$ as an engineering leakage budget for the LR validation attacker (a small deviation allowance under the balanced 5-way audit), and make statistical near-chance claims only on the final test evaluation with patient-level cluster bootstrap confidence intervals.
While early stopping is one implementation of this rule, our use of checkpoint selection is not test-set cherry-picking: the rule is specified before test evaluation and uses only the held-out LR-based validation audit on \balval.
We intentionally log the full trajectory to expose non-monotonic rebound and to make the selection decision auditable.


\section{Output-Based Privacy Evaluation Details}
\label{app:output_privacy}

\subsection{Output-Based Privacy: Text Attacker and Mention Rate}
Representation probing characterizes recorded-category signal in internal features, but it does not directly measure leakage from generated text.
We therefore add complementary output-based evaluations on the same balanced \testbalanced{} split.

\paragraph{Text attacker and mention rate.}
Given each generated summary $\hat{y}$, we train a 5-way race text attacker by fine-tuning Bio\_ClinicalBERT~\citep{alsentzer2019publicly} as a sequence classifier on generations from \baltrain{} and selecting the best checkpoint on \balval{} using Macro-F1.
We then evaluate attacker Accuracy and Macro-AUROC on \testbalanced.
This attacker models an attribute-inference scenario in which generated summaries are exposed to downstream analytics or monitoring, complementing prior work on demographic and protected-attribute leakage from text representations~\citep{elazar2018adversarial, zhang2020hurtful}.
To reduce the possibility that the attacker succeeds only by exploiting explicit demographic mentions, we additionally compute a mention rate on $\hat{y}$ using (i) strict patterns and (ii) loose patterns that also match contextual mentions.

\paragraph{Uncertainty.}
For output-based metrics, we compute cluster bootstrap confidence intervals by resampling patients (\texttt{subject\_id}) with replacement.
In the main tables, we report point estimates for readability; per-metric bootstrap summaries are saved alongside the results.

\subsection{Full Results: Bio\_ClinicalBERT Attacker and Clinical Proxies}
\label{app:output_text_and_clinical_full}

\begin{table*}[t]
\centering
\small
\setlength{\tabcolsep}{3.6pt}
\renewcommand{\arraystretch}{0.92}

\caption{Output-based privacy and clinical utility proxies on \texttt{test\_balanced}.
\textbf{SLoRA} denotes \textbf{SurfaceLoRA}.
\textbf{Text attacker:} a Bio\_ClinicalBERT classifier fine-tuned on generated summaries from \texttt{bal\_train} and selected on \texttt{bal\_val}; chance for balanced 5-way classification is 0.20.
\textbf{Clinical proxies:} BERTScore-F1 uses Bio\_Discharge\_Summary\_BERT (no-IDF); Concept-F1 uses i2b2-2010 NER (Problem/Test/Treatment) via normalized concept-set overlap between reference and generation.
\textbf{Bold} indicates the validation-selected checkpoint.
Mention-rate analyses are reported separately in Appendix~\ref{app:mention_breakdown} to avoid conflating race-group terms with meta-terms (e.g., ``race'', ``ethnicity'') induced by prompts.}
\label{tab:output_text_and_clinical_full}

\begin{tabular}{lcc|cccc}
\toprule
\textbf{Method} &
\multicolumn{2}{c|}{\textbf{Text attacker}} &
\multicolumn{4}{c}{\textbf{Clinical utility proxies}} \\
\cmidrule(lr){2-3}\cmidrule(lr){4-7}
& \textbf{Acc} & \textbf{Macro-AUROC} & \textbf{BERTScore-F1} & \textbf{Concept-F1(All)} & \textbf{Problem-F1} & \textbf{Treatment-F1} \\
\midrule
NEUTRAL prompt & 0.306 & 0.631 & 0.725 & 0.130 & 0.132 & 0.091 \\
BASE prompt    & 0.306 & 0.631 & 0.726 & 0.137 & 0.140 & 0.096 \\
\textbf{SLoRA ($\lambda{=}0.02$, s1200)} & \textbf{0.299} & \textbf{0.616} & \textbf{0.744} & \textbf{0.163} & \textbf{0.160} & \textbf{0.131} \\
SLoRA ($\lambda{=}0.02$, s2000) & 0.291 & 0.610 & 0.742 & 0.155 & 0.151 & 0.125 \\
SLoRA ($\lambda{=}0.00$, s2000) & 0.287 & 0.604 & 0.752 & 0.159 & 0.155 & 0.131 \\
SLoRA ($\lambda{=}0.20$, s600)  & 0.254 & 0.572 & 0.692 & 0.084 & 0.082 & 0.065 \\
SLoRA ($\lambda{=}0.05$, s800)  & 0.238 & 0.567 & 0.699 & 0.096 & 0.091 & 0.072 \\
SLoRA ($\lambda{=}0.05$, s2000) & 0.233 & 0.557 & 0.674 & 0.057 & 0.048 & 0.053 \\
SLoRA ($\lambda{=}0.10$, s2000) & 0.207 & 0.529 & 0.643 & 0.036 & 0.026 & 0.036 \\
SLoRA ($\lambda{=}0.20$, s2000) & 0.213 & 0.527 & 0.631 & 0.030 & 0.010 & 0.032 \\
SLoRA ($\lambda{=}0.50$, s2000) & 0.204 & 0.505 & 0.515 & 0.000 & 0.000 & 0.000 \\
\bottomrule
\end{tabular}
\end{table*}


\section{Baselines and Statistical Methods}
\label{app:baselines_stats}

\subsection{Prompt Engineering Baselines (No Fine-Tuning)}
To test whether prompting alone can mitigate recorded-attribute recoverability, we evaluate two prompt-only baselines using the same base instruction-tuned model (no LoRA, no GRL training):
BASE, a standard clinical summarization system instruction, and
NEUTRAL, which adds an explicit directive to remain neutral and avoid including race-related cues.
We use the same probing protocol as \textsc{SurfaceLoRA}: render each source clinical context into $x_{\mathrm{prompt}}$, extract prompt representations $f_{\theta}(x_{\mathrm{prompt}})$ from a prompt-only forward pass, fit the LR and MLP post-hoc probes on \baltrain, and evaluate ProbeAcc and LeakageGap on \testbalanced.
For utility, we report ROUGE on \testbalanced.

\subsection{Post-hoc Representation Sanitization Baselines (No Fine-Tuning)}
\label{sec:posthoc_baselines_appendix}

In addition to prompt engineering, we evaluate post-hoc representation sanitization methods that operate on cached prompt vectors without modifying the generator.
We focus on the exported representation \texttt{lasttok\_L-1} and apply each transform to extracted prompt vectors from \baltrain{}, \balval{}, and \testbalanced{}.

\paragraph{Baselines.}
We include four representative post-hoc methods and a no-removal reference:
(i) PCA removal (\texttt{pca\_removal\_topk});
(ii) random subspace removal (\texttt{random\_removal\_topk});
(iii) one-shot linear removal (\texttt{linear\_removal\_oneshot});
(iv) INLP~\citep{ravfogel2020null};
and (v) no removal (\texttt{no\_removal}).
All post-hoc transforms are learned on \baltrain{} and applied to \balval{} and \testbalanced{} without accessing test labels.

\paragraph{Evaluation protocol.}
For each transformed representation, we train the same linear probe and nonlinear probe on \baltrain{} and evaluate accuracy and LeakageGap on \balval{} and \testbalanced{}.
Because these baselines do not affect generation, they isolate how much leakage can be reduced purely by post-processing the exported representation.

\begin{table}[t]
\centering
\scriptsize
\setlength{\tabcolsep}{2.2pt}
\renewcommand{\arraystretch}{0.95}

\begin{tabularx}{\columnwidth}{@{}>{\raggedright\arraybackslash}X r r r r@{}}
\toprule
\textbf{Method} & \textbf{LR} & \textbf{Gap} & \textbf{MLP} & \textbf{Gap} \\
\midrule

\multicolumn{5}{@{}p{\columnwidth}@{}}{\textit{Post-hoc transforms (trained on \baltrain; no fine-tuning):}} \\
None (no rm)                 & 0.2140 & 0.0140 & 0.2156 & 0.0156 \\
PCA-rm (top-$k$)             & \textbf{0.2048} & \textbf{0.0048} & 0.2120 & 0.0120 \\
Rand-rm (top-$k$)            & 0.2144 & 0.0144 & 0.2176 & 0.0176 \\
Lin-rm (1-shot)              & 0.2136 & 0.0136 & 0.2124 & 0.0124 \\
INLP \citep{ravfogel2020null}& 0.2116 & 0.0116 & 0.2136 & 0.0136 \\
\midrule

\multicolumn{5}{@{}p{\columnwidth}@{}}{\textit{Training-time regularization (fine-tuned; probed on same representation):}} \\
XCov (ft; $\beta{=}0.5$; s1600) & 0.2312 & 0.0312 & 0.2350 & 0.0350 \\
\bottomrule
\end{tabularx}

\caption{Post-hoc and training-time leakage mitigation baselines on \testbalanced\ for \texttt{lasttok\_L-1} under balanced five-way race prediction (chance $=0.20$).
Gap (\textbf{LeakageGap}) is $|\text{Acc}-0.2|$ (lower is better).
Post-hoc transforms are learned on \baltrain\ and applied to \testbalanced\ without accessing test labels.
\textit{Abbrev.} LR: logistic-regression probe; MLP: multi-layer perceptron probe; rm: removal; ft: fine-tuned; s1600: step 1600; Rand: random; Lin: linear; PCA: principal component analysis.}
\label{tab:posthoc_lasttok}
\end{table}

\paragraph{Matched-dimension protocol and fixed hyperparameters.}
To ensure comparability across post-hoc baselines, we match removal strength by aligning the removed dimensionality.
We first run INLP on \baltrain{} representations and define the total removed dimensionality as
$k := \sum_{t=1}^{T} r_t$,
where $r_t$ is the rank of the row space of the $t$-th classifier weight matrix (estimated by SVD).
We then set PCA and random removal to remove exactly $k$ dimensions (top-$k$ principal components for PCA; a random $k$-dimensional orthonormal subspace for random removal), and apply the resulting linear projection to \balval{} and \testbalanced{}.

INLP is run for at most $T=20$ iterations.
At each iteration, we fit a 5-way multinomial logistic regression classifier on \baltrain{} with fixed hyperparameters:
\texttt{solver=saga}, \texttt{penalty=l2}, \texttt{multi\_class=multinomial}, \texttt{max\_iter=800}, \texttt{tol=1e-3}, and \texttt{n\_jobs=16} (with a fixed random seed).
Given the classifier weight matrix $W_t$, we compute its SVD and define $r_t$ as the number of singular values greater than $10^{-6}$; we then remove the corresponding row-space directions by projecting to the orthogonal complement and repeat on the projected representations.
We use \balval{} to monitor leakage via $\mathrm{LeakageGap}=|\mathrm{Acc}-0.2|$ (5-way chance is $0.2$), and early-stop INLP if $\mathrm{LeakageGap}\le 0.003$; otherwise we run the full 20 iterations.
As a boundary-case fallback, if INLP removes zero dimensions ($k=0$), we set $k$ to the rank of the one-shot linear removal classifier.

In our main setting, INLP does not trigger early stopping and runs the full 20 iterations, removing $k=100$ dimensions; therefore PCA and random removal use $k=100$.
The one-shot linear removal removes 5 dimensions in this setting.

\subsection{Statistical Uncertainty and Significance}
\label{sec:statistical_methods_appendix}

To quantify uncertainty on the fixed \testbalanced{} split (2{,}500 examples; 500 per race), we attach confidence intervals (CIs) to both utility (ROUGE) and leakage (probe accuracy).

\paragraph{ROUGE confidence intervals.}
For ROUGE, we use a paired bootstrap protocol~\citep{koehn2004statistical, tibshirani1993introduction}.
After caching one deterministic generation per example (greedy decoding; no sampling), we compute per-example ROUGE-1/2/L F1 scores by comparing each hypothesis to its reference.
We draw $B{=}10{,}000$ bootstrap resamples of size $N{=}2{,}500$ with replacement from the test indices.
For each resample, we recompute corpus-level ROUGE as the mean of per-example ROUGE scores in the resampled set.
We report 95\% CIs using the percentile method.
For pairwise comparisons, we use paired bootstrap and treat a difference as significant at $\alpha{=}0.05$ if the 95\% CI of the difference excludes zero.

\paragraph{Probe accuracy confidence intervals.}
For probing-based leakage metrics, we report probe accuracy (ProbeAcc) on the fixed \testbalanced split.
Because multiple examples may originate from the same patient (\texttt{subject\_id}), example-level binomial/Wilson intervals can be anti-conservative.
We therefore compute 95\% CIs using a patient-level stratified cluster bootstrap on \texttt{subject\_id}, preserving the balanced 5-way composition.

Concretely, let $\mathcal{I}_c$ denote the set of indices in \testbalanced{} with race label $c \in \{1,\dots,5\}$, and let $\mathcal{S}_c$ be the set of unique \texttt{subject\_id}s appearing in $\mathcal{I}_c$.
For each bootstrap replicate ($B{=}10{,}000$), we construct a balanced resample by:
(i) sampling 500 \texttt{subject\_id}s with replacement from $\mathcal{S}_c$ for each class $c$;
(ii) for each sampled patient, uniformly sampling one index from that patient’s indices in $\mathcal{I}_c$.
Concatenating across the five classes yields a bootstrap sample of size 2{,}500 (500 per class), on which we compute ProbeAcc.
We report 95\% CIs using the percentile method (2.5/97.5 percentiles over the $B$ bootstrap ProbeAcc values).

For pairwise comparisons of ProbeAcc between two methods, we use a paired patient-level bootstrap: within each replicate we reuse the same sampled patients and within-patient sampled indices for both methods. We treat a difference as significant at $\alpha{=}0.05$ if the 95\% CI of the accuracy difference excludes zero.
We report ProbeAcc as consistent with random guessing when the chance level ($0.20$) falls within the patient-level bootstrap 95\% CI.


\section{Additional Output-Based Leakage Analyses}
\label{app:output_leakage}

This appendix reports additional evaluations that quantify generated-text leakage for the case-study sensitive label, EHR-recorded race, as opposed to internal representation leakage.
All results are computed on the balanced \texttt{test\_balanced} split ($n{=}2500$; 500 per race group), with patient-level (cluster) bootstrap confidence intervals obtained by resampling \texttt{subject\_id} with replacement.

\subsection{Generated-Text Leakage Attacker: Term Frequency--Inverse Document Frequency (TF-IDF) + Logistic Regression}
\label{app:tfidf_attacker}

\paragraph{Attacker setup.}
We train a bag-of-words attacker on generated summaries using TF-IDF features with 1--2 grams and a multinomial logistic regression classifier.
The attacker is trained on \baltrain{} and tuned on \balval{} by selecting the regularization strength $C$ that maximizes validation Macro-F1 (tie-broken by Accuracy).
We then evaluate on \testbalanced{} and report Accuracy, Macro-F1, and Macro-AUROC (one-vs-rest).

\paragraph{Why include a simple attacker?}
Compared with neural attackers that may exploit deeper semantic correlates, TF-IDF+LR provides a transparent baseline that primarily captures lexical and short-context cues.
Strong performance from this model indicates that race-associated information is recoverable from relatively surface-level lexical signals.

\paragraph{Results.}
Table~\ref{tab:gen_text_leakage} summarizes leakage from generated text.
The TF-IDF attacker yields slightly different absolute accuracies from the Bio\_ClinicalBERT attacker in Table~\ref{tab:output_text_and_clinical_full}, but the qualitative conclusion is the same: generated summaries remain above chance, and representation-level mitigation does not eliminate output-level demographic predictability.
The attacker remains consistently above chance (0.20) across methods, with prompt-only baselines exhibiting the largest leakage.
SurfaceLoRA reduces leakage relative to prompt-only baselines, although it does not reduce the attacker to chance.

\begin{table*}[t]
\centering
\small
\setlength{\tabcolsep}{3.6pt}
\renewcommand{\arraystretch}{0.92}

\caption{Generated-text leakage on \texttt{test\_balanced} ($n{=}2500$; balanced).
Attacker is TF-IDF + Logistic Regression trained on \texttt{bal\_train} and tuned on \texttt{bal\_val}.
Confidence intervals are patient-level cluster bootstrap (95\%).
SLoRA denotes SurfaceLoRA; $sXXXX$ indicates the training step.
Men(S/L) are legacy mention-rate metrics (Strict/Loose; prior to the meta/group decomposition in Table~\ref{tab:mention_breakdown}).
\textbf{Bold} indicates the validation-selected checkpoint.
Chance is 0.20.}
\label{tab:gen_text_leakage}

\begin{tabular}{lcccccc}
\toprule
Method & Acc & CI Acc (95\%) & Macro-F1 & CI F1 (95\%) & Macro-AUROC & Men(S/L)\% \\
\midrule
BASE
& 0.3352 & [0.315, 0.355]
& 0.3306 & [0.310, 0.350]
& 0.6535 & 0.76 / 0.88 \\
NEUTRAL
& 0.3308 & [0.311, 0.352]
& 0.3270 & [0.307, 0.348]
& 0.6576 & 46.52 / 46.56 \\
\midrule
SLoRA $\lambda{=}0.00$, s2000
& 0.3044 & [0.284, 0.324]
& 0.2998 & [0.280, 0.318]
& 0.6270 & 0.32 / 0.40 \\
\textbf{SLoRA $\lambda{=}0.02$, s1200}
& \textbf{0.3100} & \textbf{[0.288, 0.331]}
& \textbf{0.3055} & \textbf{[0.285, 0.325]}
& \textbf{0.6330} & \textbf{0.36 / 0.48} \\
SLoRA $\lambda{=}0.02$, s2000
& 0.3032 & [0.282, 0.323]
& 0.2995 & [0.279, 0.319]
& 0.6351 & 0.36 / 0.44 \\
SLoRA $\lambda{=}0.05$, s800
& 0.2808 & [0.261, 0.303]
& 0.2776 & [0.257, 0.299]
& 0.6023 & 0.32 / 0.40 \\
SLoRA $\lambda{=}0.05$, s2000
& 0.2768 & [0.258, 0.295]
& 0.2731 & [0.254, 0.290]
& 0.6014 & 0.36 / 0.44 \\
SLoRA $\lambda{=}0.10$, s2000
& 0.2784 & [0.260, 0.297]
& 0.2761 & [0.257, 0.294]
& 0.5977 & 0.32 / 0.40 \\
SLoRA $\lambda{=}0.20$, s600
& 0.2876 & [0.269, 0.306]
& 0.2862 & [0.268, 0.304]
& 0.5946 & 0.32 / 0.44 \\
SLoRA $\lambda{=}0.20$, s2000
& 0.2868 & [0.270, 0.306]
& 0.2844 & [0.267, 0.303]
& 0.5989 & 0.44 / 0.52 \\
SLoRA $\lambda{=}0.50$, s2000
& 0.2976 & [0.277, 0.318]
& 0.2943 & [0.274, 0.313]
& 0.6229 & 0.32 / 0.40 \\
\bottomrule
\end{tabular}
\end{table*}

\subsection{Reference Attacker as a Diagnostic Baseline}
\label{app:reference_upper_bound}

To contextualize output leakage, we also train the same TF-IDF+LR attacker on the gold reference targets (BHC) as a diagnostic baseline for demographic inference from the narrative itself.
On \texttt{test\_balanced}, the reference attacker achieves
Acc$=0.3284$, Macro-F1$=0.3255$, and Macro-AUROC$=0.6566$.
These results indicate that correlates of the case-study sensitive label are present even in the ground-truth clinical narrative.
Accordingly, output-based inference of EHR-recorded race may persist even when models avoid explicit mentions, and stronger attackers could potentially achieve higher performance than this baseline.

\subsection{Decomposed Mention Rate: Meta Terms vs.\ Group Terms}
\label{app:mention_breakdown}

\paragraph{Motivation and definitions.}
A naive mention-rate metric can be confounded by meta terms (e.g., ``race'', ``ethnicity'', ``demographics'') that appear due to prompt wording rather than demographic disclosure.
We therefore decompose mention signals into:
(i) meta-term rate (generic references to race/ethnicity) and
(ii) group-term rate (explicit group identifiers such as ``Hispanic'' or ``Asian'').
For ambiguous terms such as \textit{black}/\textit{white}, we apply a strict local-context filter to reduce false positives (e.g., ``black stool'', ``white blood cell count'') and count these tokens only when identity cues (e.g., ``patient'', ``race'', ``ethnicity'') are present and medical-color cues are absent.

\paragraph{Results on \texttt{test\_balanced}.}
Table~\ref{tab:mention_breakdown} reports decomposed mention rates.
The neutral prompt baseline exhibits an extremely high meta-term rate (46.44\%) driven by prompt-induced wording, while its group-term rate remains low (0.48\%).
In contrast, other methods have very low meta-term rates ($\leq 0.12\%$) and low group-term rates ($\leq 0.96\%$), suggesting that text-attacker leakage is not primarily driven by overt demographic mentions.

\begin{table*}[t]
\centering
\small
\setlength{\tabcolsep}{6.0pt}
\caption{Decomposed mention rate on \texttt{test\_balanced} ($n{=}2500$). Meta counts generic references to race/ethnicity (e.g., \textit{race}, \textit{racial}, \textit{ethnicity}, \textit{demographics}). Group counts explicit group identifiers (e.g., \textit{Hispanic}, \textit{Asian}) with strict context filtering for \textit{black}/\textit{white}. Meta-only and Group-only exclude overlaps. \textbf{Bold} indicates the validation-selected checkpoint.}
\label{tab:mention_breakdown}
\begin{tabular}{lccccc}
\toprule
Method & Group(\%) & Meta(\%) & Meta-only(\%) & Group-only(\%) & Both(\%) \\
\midrule
prompt\_base
& 0.96 & 0.08 & 0.08 & 0.96 & 0.00 \\
prompt\_neutral
& 0.48 & 46.44 & 46.16 & 0.20 & 0.28 \\
\midrule
SurfaceLoRA\_lam0.00\_step2000
& 0.48 & 0.04 & 0.04 & 0.48 & 0.00 \\
\textbf{SurfaceLoRA\_lam0.02\_step1200}
& \textbf{0.56} & \textbf{0.04} & \textbf{0.04} & \textbf{0.56} & \textbf{0.00} \\
SurfaceLoRA\_lam0.02\_step2000
& 0.48 & 0.04 & 0.04 & 0.48 & 0.00 \\
SurfaceLoRA\_lam0.05\_step800
& 0.48 & 0.04 & 0.04 & 0.48 & 0.00 \\
SurfaceLoRA\_lam0.05\_step2000
& 0.44 & 0.08 & 0.08 & 0.44 & 0.00 \\
SurfaceLoRA\_lam0.10\_step2000
& 0.44 & 0.04 & 0.04 & 0.44 & 0.00 \\
SurfaceLoRA\_lam0.20\_step600
& 0.48 & 0.04 & 0.04 & 0.48 & 0.00 \\
SurfaceLoRA\_lam0.20\_step2000
& 0.52 & 0.12 & 0.08 & 0.48 & 0.04 \\
SurfaceLoRA\_lam0.50\_step2000
& 0.44 & 0.04 & 0.04 & 0.44 & 0.00 \\
\bottomrule
\end{tabular}
\end{table*}

\subsection{Takeaways}
\label{app:output_takeaways}

Across methods, the TF-IDF+LR attacker remains above chance on generated text, indicating that demographic inference can be feasible from relatively surface lexical cues even when explicit demographic mentions are rare.
The decomposed mention-rate analysis supports this interpretation: for most methods, overt race-group identifiers occur infrequently, so residual output leakage likely arises from implicit correlates in clinical narratives (e.g., comorbidities, social history, medications, or care patterns) rather than direct demographic statements.
Finally, the reference attacker trained on gold BHC targets suggests that race-associated correlates are present in the underlying narrative distribution, providing context for why output-based inference may persist even when representation-level leakage is reduced.



\section{Additional Validation and Group-Wise Results}

\subsection{Validation-Time Pareto Analysis and Best Checkpoints}
\label{app:pareto_val_points}

During training, we evaluate every 200 steps and record validation ROUGE and LR probe accuracy, where the LR probe is fit on \baltrain\ and evaluated on \balval.
For deployment selection, we apply the checkpoint selection rule in Appendix~\ref{app:checkpoint_governance}: choose the highest-ROUGE-L checkpoint among those satisfying the LR-based validation leakage budget.
For diagnostic reporting, we also compute the Pareto front under minimizing LR \texttt{LeakageGap} and maximizing ROUGE-L.
The validation-selected checkpoint and Pareto-optimal validation points are summarized in Table~\ref{tab:pareto_val_points}.

\begin{table}[t]
\centering
\small
\setlength{\tabcolsep}{5pt}
\begin{tabular}{@{}lcccccc@{}}
\toprule
Run & Step & R-1 & R-2 & R-L & LR Acc & LR Gap \\
\midrule
$\lambda{=}0.20$ & 600  & 22.53 & 5.15 & 12.14 & 0.217 & 0.017 \\
$\lambda{=}0.50$ & 600  & 17.29 & 3.75 & 9.18  & 0.218 & 0.018 \\
$\lambda{=}0.05$ & 800  & 23.72 & 5.54 & 12.87 & 0.222 & 0.022 \\
\textbf{$\lambda{=}0.02$} & \textbf{1200} & \textbf{27.81} & \textbf{7.45} & \textbf{14.40} & \textbf{0.223} & \textbf{0.023} \\
$\lambda{=}0.10$ & 400  & 25.39 & 6.37 & 13.45 & 0.225 & 0.025 \\
$\lambda{=}0.00$ & 200  & 25.07 & 6.07 & 13.46 & 0.229 & 0.029 \\
\midrule
\multicolumn{7}{@{}l@{}}{Pareto front (subset of rows above):} \\
$\lambda{=}0.20$ & 600  & 22.53 & 5.15 & 12.14 & 0.217 & 0.017 \\
$\lambda{=}0.05$ & 800  & 23.72 & 5.54 & 12.87 & 0.222 & 0.022 \\
\textbf{$\lambda{=}0.02$} & \textbf{1200} & \textbf{27.81} & \textbf{7.45} & \textbf{14.40} & \textbf{0.223} & \textbf{0.023} \\
\bottomrule
\end{tabular}
\caption{Validation checkpoints and Pareto front for the \texttt{lasttok\_L-1} representation.
R-1/R-2/R-L are ROUGE-1/2/L, LR Acc is logistic-regression probe accuracy, and LR Gap is $|\text{LR Acc}-0.2|$.
The validation LR probe is fit on \baltrain\ and evaluated on \balval.
\textbf{Bold} indicates the validation-selected checkpoint chosen by maximizing validation ROUGE-L subject to the LR-based validation leakage budget.}
\label{tab:pareto_val_points}
\end{table}

\subsection{Meanpool-Targeted Validation Pareto Analysis}
\label{app:pareto_val_points_meanpool}

We repeat the same checkpoint selection procedure using the \texttt{meanpool\_L-1} representation, where mean pooling is computed over all non-padding tokens in $x_{\mathrm{prompt}}$.
Across evaluated checkpoints, no model satisfies the LR-based validation leakage budget for this representation; indeed, all evaluated meanpool checkpoints remain well above chance (Gap $>0.10$).
We therefore report Pareto-optimal checkpoints and select the minimum-gap point as the fallback checkpoint.
Table~\ref{tab:meanpool_best_per_lambda} summarizes the best diagnostic checkpoint for each $\lambda$ after the LR-based leakage-budget rule fails for \texttt{meanpool\_L-1}; the bold row marks the fallback checkpoint with the smallest validation LeakageGap.
Table~\ref{tab:pareto_val_points_meanpool} reports the validation-time Pareto front under minimizing Gap and maximizing ROUGE-L.

\begin{table}[t]
\centering
\small
\setlength{\tabcolsep}{5pt}
\begin{tabular}{@{}r r r r r@{}}
\toprule
$\lambda$ & Step & R-L & LR Acc & LR Gap \\
\midrule
0.00 & 800  & 13.67 & 0.3164 & 0.1164 \\
\textbf{0.05} & \textbf{400}  & \textbf{13.21} & \textbf{0.3084} & \textbf{0.1084} \\
0.10 & 400  & 12.97 & 0.3132 & 0.1132 \\
0.30 & 600  & 13.91 & 0.3136 & 0.1136 \\
1.00 & 800  & 13.53 & 0.3136 & 0.1136 \\
\bottomrule
\end{tabular}
\caption{Diagnostic validation checkpoints per $\lambda$ under the \texttt{meanpool\_L-1} representation, computed from $x_{\mathrm{prompt}}$.
Because no evaluated meanpool checkpoint satisfies the leakage budget, rows are shown for diagnostic comparison and the \textbf{bold} row indicates the fallback checkpoint with minimum validation LeakageGap.}
\label{tab:meanpool_best_per_lambda}
\end{table}

\begin{table}[t]
\centering
\small
\setlength{\tabcolsep}{5pt}
\begin{tabular}{@{}r r r r r@{}}
\toprule
$\lambda$ & Step & R-L & LR Acc & LR Gap \\
\midrule
\textbf{0.05} & \textbf{400}  & \textbf{13.21} & \textbf{0.3084} & \textbf{0.1084} \\
0.05 & 1000 & 14.24 & 0.3132 & 0.1132 \\
0.05 & 1200 & 14.65 & 0.3136 & 0.1136 \\
0.30 & 1200 & 14.66 & 0.3212 & 0.1212 \\
1.00 & 1400 & 14.79 & 0.3260 & 0.1260 \\
\bottomrule
\end{tabular}
\caption{Validation-time Pareto front under the \texttt{meanpool\_L-1} representation, computed from $x_{\mathrm{prompt}}$, optimizing lower validation LeakageGap and higher validation ROUGE-L.
Because no checkpoint satisfies the leakage budget, the \textbf{bold} row indicates the fallback checkpoint with minimum validation LeakageGap.}
\label{tab:pareto_val_points_meanpool}
\end{table}

\paragraph{Takeaway.}
Compared with \texttt{lasttok}-targeted training, meanpool-targeted runs do not reach chance-level recovery under the validation audit: the best observed Gap is $0.108$ (ProbeAcc $=0.308$).
This reinforces the deployment rule that mitigation should be evaluated on the exact exported vector, and that mean-pooled prompt embeddings constitute a distinct and harder-to-sanitize artifact if used in deployment.

\subsection{Group-Wise Utility Stability Analysis}
\label{app:groupwise_stability}

To assess whether leakage reduction comes at the cost of subgroup-specific utility degradation, we evaluate group-wise ROUGE-L on the balanced \testbalanced{} split (500 examples per race group).
We report per-group ROUGE-L means, the absolute gap (max $-$ min), the standard deviation across groups, worst-group performance, and the relative gap (gap divided by the overall mean).

\begin{table*}[t]
\centering
\small
\begin{tabular}{lcccccccccc}
\toprule
Method & Overall & White & Black & Hisp. & Asian & Other & Gap$\downarrow$ & Std$\downarrow$ & Worst$\uparrow$ & RelGap$\downarrow$ \\
\midrule
\multicolumn{11}{l}{Validation-screened early checkpoints (selected on \balval; evaluated on \testbalanced; bold row is selected):} \\
\textbf{$\lambda{=}0.02$, step 1200} & \textbf{14.51} & \textbf{14.46} & \textbf{14.31} & \textbf{14.37} & \textbf{14.68} & \textbf{14.71} & \textbf{0.40} & \textbf{0.16} & \textbf{14.31} & \textbf{0.028} \\
$\lambda{=}0.20$, step 600  & 12.42 & 12.60 & 12.33 & 12.05 & 12.52 & 12.62 & 0.58 & 0.21 & 12.05 & 0.046 \\
$\lambda{=}0.05$, step 800  & 12.89 & 12.97 & 12.71 & 12.65 & 13.10 & 13.01 & 0.45 & 0.18 & 12.65 & 0.035 \\
\midrule
\multicolumn{11}{l}{Full-run checkpoints (step 2{,}000; no early stopping):} \\
$\lambda{=}0.00$ & 15.02 & 15.21 & 14.80 & 15.00 & 14.88 & 15.22 & 0.42 & 0.17 & 14.80 & 0.028 \\
$\lambda{=}0.02$ & 14.38 & 14.52 & 14.11 & 14.56 & 14.29 & 14.43 & 0.46 & 0.17 & 14.11 & 0.032 \\
$\lambda{=}0.05$ & 11.69 & 11.73 & 11.83 & 11.47 & 11.40 & 12.05 & 0.65 & 0.24 & 11.40 & 0.055 \\
$\lambda{=}0.10$ & 10.43 & 10.31 & 10.23 & 10.20 & 10.82 & 10.61 & 0.62 & 0.24 & 10.20 & 0.059 \\
$\lambda{=}0.20$ & 9.04 & 8.99 & 8.81 & 9.08 & 9.20 & 9.10 & 0.39 & 0.13 & 8.81 & 0.043 \\
$\lambda{=}0.50$ & 7.28 & 7.08 & 7.04 & 7.17 & 7.63 & 7.46 & 0.59 & 0.23 & 7.04 & 0.082 \\
\midrule
\multicolumn{11}{l}{Prompt engineering baselines (no training):} \\
NEUTRAL prompt & 12.81 & 12.82 & 12.64 & 12.78 & 12.84 & 12.95 & 0.31 & 0.10 & 12.64 & 0.025 \\
BASE prompt    & 12.87 & 12.88 & 12.62 & 12.88 & 12.93 & 13.04 & 0.42 & 0.14 & 12.62 & 0.033 \\
\bottomrule
\end{tabular}
\caption{Group-wise ROUGE-L on the balanced \testbalanced{} split (500 examples per race group; $n{=}2500$).
Gap = max $-$ min across groups; Std = population standard deviation across groups; Worst = minimum group performance; RelGap = Gap / Overall.
Lower Gap, Std, and RelGap indicate more consistent utility across groups; higher Worst indicates better worst-group utility.
\textbf{Bold} indicates the validation-selected checkpoint.}
\label{tab:groupwise_stability}
\end{table*}

\paragraph{Group-wise utility remains stable.}
The selected checkpoint ($\lambda{=}0.02$, step 1200) achieves overall ROUGE-L 14.51 with a gap of 0.40 points (relative gap 2.8\%).
Performance across the five race groups is tightly clustered (White 14.46, Black 14.31, Hispanic 14.37, Asian 14.68, Other 14.71).
This gap is comparable to the vanilla LoRA full-run checkpoint ($\lambda{=}0.00$, step 2{,}000: gap 0.42, relative gap 2.8\%), suggesting that exported-vector-targeted adversarial training does not introduce large group-conditional differences in summarization quality.

\paragraph{Degradation under large $\lambda$ is primarily global.}
As $\lambda$ increases in full-run training, the main change is a decrease in overall ROUGE-L (e.g., 15.02 at $\lambda{=}0.00$ vs.\ 7.28 at $\lambda{=}0.50$), while absolute group gaps remain modest (0.39--0.65).
The larger relative gaps at low utility (e.g., $\lambda{=}0.50$) are driven largely by the smaller overall denominator.
Overall, these results suggest that the fragility observed in Sections~\ref{sec:fragility}--\ref{sec:test_all_methods} manifests primarily as global utility loss rather than amplified cross-group variation.

\paragraph{Prompting alone does not achieve the same trade-off.}
While NEUTRAL exhibits slightly smaller absolute group gaps than some fine-tuned models, its overall ROUGE-L is substantially lower than the best SurfaceLoRA checkpoint, and its representations remain highly probeable for race (Table~\ref{tab:test_all_methods}).
Prompt-only methods therefore do not simultaneously achieve strong utility and low representation-level leakage.

\paragraph{Summary.}
Across all evaluated methods, group-wise ROUGE-L gaps remain small in absolute terms ($\leq 0.65$ points) and comparable to non-adversarial baselines.
SurfaceLoRA reduces representational EHR-recorded race recoverability (Table~\ref{tab:test_all_methods}) without increasing cross-group utility variation, supporting its use as a practical leakage-mitigation approach in clinical summarization.

\section{Additional Attribute Stress Test: EHR-Recorded Gender} 
\label{app:gender_stress}

To assess whether the representation-specific pattern is limited to EHR-recorded race, we run an additional stress test using binary EHR-recorded gender.
As with EHR-recorded race, this variable should be interpreted as a recorded administrative field rather than a complete account of sex or gender identity~\citep{heidari2018sex}.
This experiment uses the same exported representations and the same post-hoc logistic-regression probing protocol used in the main audit, but replaces the training-time adversarial discriminator with a stronger two-layer MLP discriminator.
Because the audit uses a balanced binary gender subset, the chance baseline is $0.50$; these results should not be interpreted as chance-level gender sanitization.
Instead, they test whether mitigation remains artifact-specific under another recorded attribute.

Table~\ref{tab:gender_stress} shows that gender recoverability from the targeted \texttt{lasttok} artifact decreases under adversarial training, while generation utility remains comparable.
For example, the best final-step MLP-discriminator run reduces \texttt{lasttok} probe accuracy from $0.769$ to $0.700$ with similar ROUGE-L ($15.516$ vs.\ $15.417$).
However, \texttt{lasttok} recoverability remains well above the $0.50$ chance baseline.
At the same time, \texttt{meanpool} remains near-saturated ($0.995$--$0.997$), producing a large gap across exported artifacts.
Thus, the gender stress test supports the main finding that leakage and mitigation are representation-specific, while also showing that mitigation difficulty is attribute-dependent.

\begin{table}[t]
\centering
\scriptsize
\setlength{\tabcolsep}{3.6pt}
\renewcommand{\arraystretch}{0.95}
\begin{tabular}{lccccc}
\toprule
\textbf{Setting} & \textbf{R-L} & \textbf{Lasttok} & \textbf{Gap} & \textbf{Meanpool} & $\boldsymbol{\Delta}$ \\
\midrule
No mitigation, $\lambda{=}0.000$ & 15.516 & 0.769 & 0.269 & 0.997 & 0.228 \\
Val-selected, $\lambda{=}0.015$ & 15.131 & 0.702 & 0.202 & 0.995 & 0.293 \\
Best final, $\lambda{=}0.030$ & 15.417 & 0.700 & 0.200 & 0.995 & 0.295 \\
\bottomrule
\end{tabular}
\caption{Additional EHR-recorded gender stress test using the same exported-representation audit.
The training-time adversarial discriminator is a two-layer MLP, while evaluation uses the same post-hoc logistic-regression probing protocol as the main experiments.
Chance accuracy for the balanced binary gender audit is $0.50$.
Gap is $|\text{Lasttok Acc}-0.50|$, and $\Delta$ is \texttt{meanpool} accuracy minus \texttt{lasttok} accuracy.
The validation-selected row uses $\lambda{=}0.015$ at step 1600; the best final-step row uses $\lambda{=}0.030$ at the final training step.
These results are not evidence of chance-level gender sanitization; rather, they show that even when recoverability decreases on the targeted \texttt{lasttok} artifact, an untargeted pooled representation can remain near-saturated.}
\label{tab:gender_stress}
\end{table}


\section{Additional Discussion: Representation Mismatch and Audit Rule}

\subsection{System Warning: Representation Mismatch and the Need to Audit the Exported Artifact}
\label{app:audit_the_artifact_rule}

Our representation-sensitivity results (Section~\ref{sec:redistribution}) highlight a deployment pitfall: mitigation on one exported vector may not transfer to another.
A model can appear mitigated at one representation choice (e.g., the last-token prompt vector) while remaining strongly vulnerable to attribute inference at another (e.g., mean-pooled embeddings).
This indicates that leakage reduction is not generally transferable across aggregation strategies.
Consequently, model-level claims about mitigation can create a false sense of security if the deployment pipeline exports a representation that differs from the artifact targeted during training.

\paragraph{Audit rule: audit the artifact.}
Deployers should not only ask whether a model is ``leakage-mitigated,'' but whether the specific exported representation (the artifact that is stored, logged, indexed, cached, or reused) is mitigated under the attacker classes of interest.
If a deployment exports mean-pooled embeddings, common in retrieval pipelines, the adversarial objective should be attached to that representation; sanitizing the last-token representation alone is insufficient.



\section{Implementation Details}

\paragraph{Prompt construction (exact template).}
We construct prompts using the base model's tokenizer-defined chat template rather than a hand-written string format.
For each example, let $c$ denote the source clinical context.
In the primary dataset, $c$ is the discharge-note input with the BHC section removed.
We create (i) a system message (``You are a clinical assistant. Summarize the note into a Brief Hospital Course.'') and (ii) a user message containing $c$.
We then render these messages with \texttt{tokenizer.apply\_chat\_template(..., add\_generation\_prompt=True)}, which appends the assistant generation header and injects model-specific special tokens and role delimiters.
The resulting rendered prompt is denoted $x_{\mathrm{prompt}}$.
Because hidden states depend on these special tokens, using the tokenizer-provided template is important for reproducibility.

\paragraph{Padding and truncation.}
We use left padding for batching.
If \texttt{pad\_token\_id} is undefined, we set \texttt{pad\_token=eos\_token}.
For prompt-only forward passes used in probing and adversarial training, we truncate $x_{\mathrm{prompt}}$ to at most 1{,}024 tokens by keeping the most recent tokens.
For SFT, we concatenate $x_{\mathrm{prompt}}$ with the target BHC tokens and truncate the full prompt--target sequence to at most 1{,}536 tokens using the same rule.

\paragraph{Representation for adversary and probing.}
Let $f_{\theta}(x_{\mathrm{prompt}})$ denote the exported representation of the rendered prompt.
Unless otherwise stated, we run a \emph{prompt-only} forward pass on $x_{\mathrm{prompt}}$ with \texttt{output\_hidden\_states=True}.
With left padding, let $i^\star = \max\{i:\texttt{attention\_mask}_i=1\}$ be the index of the last non-padding token in $x_{\mathrm{prompt}}$, and let $\mathbf{h}^{(\ell)}_{i}(x_{\mathrm{prompt}})$ denote the hidden state at position $i$ from block $\ell$.

\noindent Default representation (\texttt{lasttok\_L-1}):
\[
f_{\theta}(x_{\mathrm{prompt}})
=
\mathbf{h}^{(L-1)}_{i^\star}(x_{\mathrm{prompt}}).
\]
\noindent Meanpool representation (\texttt{meanpool\_L-1}, when used): letting $\mathcal{P}=\{i:\texttt{attention\_mask}_i=1\}$ be the set of non-padding token positions in $x_{\mathrm{prompt}}$,
\[
f_{\theta}(x_{\mathrm{prompt}})
=
\frac{1}{|\mathcal{P}|}\sum_{i\in\mathcal{P}} \mathbf{h}^{(L-1)}_{i}(x_{\mathrm{prompt}}).
\]
For meanpool-targeted experiments, the GRL adversary and the probing classifiers are attached to this same \texttt{meanpool\_L-1} representation; otherwise they use \texttt{lasttok\_L-1}.

\paragraph{Generation and ROUGE evaluation.}
We generate summaries with greedy decoding (\texttt{do\_sample=False}) and \texttt{max\_new\_tokens=256}.
To reduce OOM risk during validation/testing, we generate in micro-batches (e.g., 8 prompts per batch) and clear the CUDA cache between batches.
ROUGE is computed offline at the word level after lowercasing and whitespace normalization.

\paragraph{MLP probe training details.}
The 2-layer MLP probe has the form:
Linear(hidden\_dim $\rightarrow$ 128) $\rightarrow$ ReLU $\rightarrow$ Dropout(0.3) $\rightarrow$ Linear(128 $\rightarrow$ 5).
We train with AdamW (learning rate $10^{-3}$, weight decay $0.01$, batch size 128) for up to 50 epochs with early stopping (patience 5 epochs) based on \balval{} accuracy.
We select the checkpoint with the best validation accuracy and evaluate it on \testbalanced.
All MLP probes are trained independently for each model checkpoint.

\paragraph{Prompt-only baselines.}
For prompt-only baselines (BASE/NEUTRAL), we keep the underlying model weights frozen (no LoRA and no adversary training) and vary only the instruction text in the system message.
We extract the same prompt representation $f_{\theta}(x_{\mathrm{prompt}})$ and evaluate it using the same LR and MLP post-hoc probing protocol as \textsc{SurfaceLoRA}.

\paragraph{Post-hoc and decorrelation baselines.}
To contextualize \textsc{SurfaceLoRA}, we additionally evaluate leakage-mitigation baselines in Appendix~\ref{sec:posthoc_baselines_appendix}.
The post-hoc baselines operate on cached \texttt{lasttok\_L-1} vectors without changing the generator: PCA removal, random subspace removal, one-shot linear removal, and INLP.
We also include XCov as a training-time decorrelation baseline, which adds a cross-covariance penalty between exported representations and one-hot race labels during fine-tuning.
All baselines are evaluated with the same balanced five-way probing protocol used for \textsc{SurfaceLoRA}.

\paragraph{Post-hoc sanitization details.}
For post-hoc baselines (Section~\ref{sec:posthoc_baselines_appendix}), we learn a linear transform on \baltrain{} representations and apply it to \balval{} and \testbalanced{}.
PCA removal projects onto the orthogonal complement of a top-$k$ PCA subspace estimated from \baltrain.
Random removal removes a matched $k$-dimensional random subspace.
One-shot linear removal applies a single supervised nullspace projection derived from a linear classifier.
INLP iteratively fits linear classifiers and composes nullspace projections as in \citet{ravfogel2020null}.
Hyperparameters (e.g., $k$ and the number of INLP iterations) are fixed across splits and determined on \balval.


\section{Additional Backbone Results: Qwen Replication}
\label{app:qwen_backbone}

This appendix reports a backbone replication on \textbf{Qwen-2.5-7B-Instruct}~\citep{bai2025qwen2} using the same dataset splits, balanced leakage subsets, probing protocol, and decoding settings as the main experiments.
We report a unified evaluation on \testbalanced{} that includes prompt-only baselines, validation-selected checkpoints, and train-to-end checkpoints.

\begin{table*}[t]
\centering
\scriptsize
\setlength{\tabcolsep}{3.5pt}
\renewcommand{\arraystretch}{1.10}
\resizebox{\linewidth}{!}{%
\begin{tabular}{llrccccc}
\toprule
Method & Checkpoint & $\mathbf{N}$ &
R-1 & R-2 & R-L &
ProbeAcc & LeakageGap \\
\midrule
\multicolumn{8}{l}{Reference (chance under balanced 5-way classification): $p=0.2$} \\
-- & chance & -- & -- & -- & -- & 0.2000 & 0.0000 \\
\midrule
\multicolumn{8}{l}{Prompt-only baselines (no fine-tuning):} \\
prompt\_only & base prompt    & 2500 & 25.0052 & 5.7580 & 12.2079 & 0.2276 & 0.0276 \\
prompt\_only & neutral prompt & 2500 & 25.4523 & 5.8731 & 12.3975 & 0.2200 & 0.0200 \\
\midrule
\multicolumn{8}{l}{Validation-screened checkpoints under the leakage-budget rule:} \\
\textbf{SurfaceLoRA} & \textbf{$\lambda=0.20$, step 1000} & \textbf{2500} & \textbf{27.2120} & \textbf{7.1622} & \textbf{13.9781} & \textbf{0.2004} & \textbf{0.0004} \\
SurfaceLoRA & $\lambda=0.02$, step 1800 & 2500 & 27.1907 & 7.0595 & 13.9354 & 0.2116 & 0.0116 \\
SurfaceLoRA & $\lambda=0.02$, step 1600 & 2500 & 27.1228 & 7.1480 & 13.7683 & 0.2144 & 0.0144 \\
\midrule
\multicolumn{8}{l}{Train-to-end checkpoints (final step=2000; no early stopping):} \\
SurfaceLoRA & $\lambda=0.00$, step 2000 & 2500 & 28.2674 & 7.5514 & 14.4109 & 0.2312 & 0.0312 \\
SurfaceLoRA & $\lambda=0.02$, step 2000 & 2500 & 24.0533 & 5.9205 & 12.4383 & 0.2100 & 0.0100 \\
SurfaceLoRA & $\lambda=0.05$, step 2000 & 2500 & 20.6192 & 4.8550 & 10.7769 & 0.2192 & 0.0192 \\
SurfaceLoRA & $\lambda=0.20$, step 2000 & 2500 & 15.9894 & 3.7770 &  8.5156 & 0.2224 & 0.0224 \\
\bottomrule
\end{tabular}%
}
\caption{Qwen replication on \texttt{test\_balanced}. LeakageGap is defined as $|\text{ProbeAcc}-0.2|$ under balanced 5-way race classification.
We report prompt-only baselines, validation-selected checkpoints, and train-to-end checkpoints.
$N$ is the number of evaluated examples; \texttt{test\_balanced} contains 2{,}500 examples.
\textbf{Bold} indicates the validation-selected checkpoint chosen by maximizing validation ROUGE-L subject to the validation LeakageGap budget on the exported artifact.}
\label{tab:qwen_unified}
\end{table*}


\section{Cross-Dataset Stress Test on Discharge Me}
\label{app:dischargeme}

\subsection{How Discharge Me differs from MIMIC-IV-Ext-BHC}
Compared to MIMIC-IV-Ext-BHC (discharge-summary input with BHC removed), Discharge Me packages inputs in an ED-centric format that can include structured fields (chief complaint/diagnosis codes) and radiology report text, and provides extracted targets (BHC and discharge instructions).
This changes the input composition and information density, and we therefore treat Discharge Me as a dataset-construction stress test of the same summarization objective.

\paragraph{Dataset overview.}
We evaluate robustness on \textit{Discharge Me}, a BioNLP@ACL'24 shared-task dataset hosted on PhysioNet (v1.3)~\citep{xu2024discharge,goldberger2000physiobank}.
Discharge Me is derived from MIMIC-IV-Note and MIMIC-IV-ED~\citep{johnson2023mimic, xu2024discharge}; each admission includes ED chief complaint and diagnosis codes, at least one radiology report, and a discharge summary with extracted targets for Brief Hospital Course (BHC) and Discharge Instructions~\citep{xu2024discharge}.
The official split sizes are train (68{,}785), validation (14{,}719), phase-I test (14{,}702), and phase-II test (10{,}962)~\citep{xu2024discharge}.

\paragraph{Positioning and interpretation.}
Because Discharge Me is also derived from MIMIC-IV sources~\citep{johnson2023mimic, xu2024discharge}, it is not intended as cross-institution generalization.
Instead, we use it as a construction/packaging stress test: relative to our primary corpus, Discharge Me packages inputs in an ED-centric form (chief complaint, ICD codes, radiology reports) and defines targets via a shared-task extraction pipeline~\citep{xu2024discharge}.
We therefore interpret results as robustness to dataset formulation and task packaging rather than fully independent external validation.

\begin{table*}[ht]
\centering
\small
\setlength{\tabcolsep}{5.2pt}
\begin{tabular}{l c c c c c c c c}
\toprule
$\lambda$ & Rep &
BestStep &
Val RL & Val Probe & Val Gap &
Test RL & Test Probe & Test Gap \\
\midrule
0.00 & lasttok & 600  & 10.9353 & 0.2222 & 0.0222 & 10.9189 & 0.2124 & 0.0124 \\
\textbf{0.02} & \textbf{lasttok} & \textbf{1600} & \textbf{11.0001} & \textbf{0.2066} & \textbf{0.0066} & \textbf{10.5743} & \textbf{0.2000} & \textbf{0.0000} \\
0.05 & lasttok & 1600 &  5.4291 & 0.2165 & 0.0165 &  6.2558 & 0.2198 & 0.0198 \\
0.10 & lasttok & 1400 &  5.9847 & 0.2099 & 0.0099 &  6.4328 & 0.2157 & 0.0157 \\
0.20 & lasttok & 2000 &  7.2862 & 0.2140 & 0.0140 &  6.9280 & 0.2298 & 0.0298 \\
\bottomrule
\end{tabular}
\caption{Discharge Me (balanced 5-way race) GRL sweep on the \texttt{lasttok} representation.
Validation metrics are reported at the selected checkpoint (BestStep) chosen by the validation selection rule: maximize validation ROUGE-L subject to the validation LeakageGap budget on the exported artifact; if no checkpoint satisfies the budget, select a Pareto-optimal checkpoint with minimum LeakageGap.
Test metrics are obtained by evaluating the same selected checkpoint on the balanced test split ($n{=}1210$).
Chance baseline for race prediction is $0.2$.}
\label{tab:dischargeme_sweep}
\end{table*}

\paragraph{Our derived ED-inputs $\rightarrow$ BHC benchmark (balanced 5-way race).}
To align with our threat model and avoid confounding from label imbalance, we construct an ED-inputs $\rightarrow$ BHC subset from Discharge Me as follows.
We join the provided tables (\texttt{edstays}, \texttt{triage}, \texttt{diagnosis}, \texttt{radiology}, \texttt{target}) using \texttt{stay\_id} and \texttt{hadm\_id}.
We form the input by concatenating: (i) chief complaint (\texttt{triage.chiefcomplaint}), (ii) ICD diagnosis codes (\texttt{diagnosis.icd\_code} with \texttt{icd\_version}), and (iii) radiology report text (\texttt{radiology.text}).
The target is \texttt{target.brief\_hospital\_course}.
We then create balanced 5-class race subsets with labels \{\textsc{ASIAN}, \textsc{BLACK}, \textsc{HISPANIC}, \textsc{OTHER}, \textsc{WHITE}\}, using equal counts per class in each split:
train $n{=}9705$ (1941 per class), validation $n{=}1215$ (243 per class), and test $n{=}1210$ (242 per class).
This balanced construction sets the chance baseline for race prediction to $1/5=0.2$, matching our leakage metric.

\paragraph{Preprocessing (provided by the dataset).}
Discharge Me provides structured source tables and extracted discharge-summary targets for the shared task, including Brief Hospital Course and Discharge Instructions sections~\citep{xu2024discharge}.
The dataset is derived from MIMIC-IV-Note and MIMIC-IV-ED, and each admission includes ED chief complaint information, diagnosis codes, at least one radiology report, and a discharge summary containing the two target sections~\citep{xu2024discharge}.
We use the provided Brief Hospital Course target field and construct the input from the shared-task source tables as described above.

\paragraph{Evaluation protocol.}
We use the same method and hyperparameters as in the main experiments unless noted otherwise.
Utility is measured by ROUGE (we report ROUGE-L), using greedy decoding with a fixed \texttt{max\_new\_tokens}.
Leakage is measured by a post-hoc linear probe (multinomial logistic regression) trained to predict race from the exported representation.
Following our threat model, the representation is the hidden state at the last non-padding prompt token (generation boundary); we denote probe accuracy as \textit{ProbeAcc} and define the deviation-from-chance metric
$\textit{Gap}=|\textit{ProbeAcc}-0.2|$.
We sweep GRL strength $\lambda \in \{0, 0.02, 0.05, 0.10, 0.20\}$ with the same representation type (\texttt{lasttok}).

\paragraph{Results and conclusions.}
Table~\ref{tab:dischargeme_sweep} reports the full $\lambda$ sweep on our balanced Discharge Me benchmark.
Across this dataset construction shift, we observe the same qualitative behavior as in our primary corpus: a small adversarial strength ($\lambda=0.02$) yields the best trade-off, substantially reducing extractable race signal on the last-token representation while maintaining competitive ROUGE-L.
In particular, $\lambda=0.02$ reaches near-chance probe performance (best validation $\textit{ProbeAcc}\approx0.2066$; final test $\textit{ProbeAcc}=0.2000$) with only a modest ROUGE-L decrease relative to $\lambda=0$.
Larger $\lambda$ values degrade utility substantially without consistently improving probe results, suggesting that over-regularization can harm generation while not guaranteeing more stable representation sanitization.

\paragraph{Access and compliance.}
Discharge Me is a credentialed-access PhysioNet dataset; we accessed the data under the PhysioNet Credentialed Health Data Use Agreement.
We did not send any dataset content to third-party APIs, consistent with the dataset rules.

\end{document}